# MFRL-BI: Design of a Model-free Reinforcement Learning Process Control Scheme by Using Bayesian Inference

Yanrong Li[1], Juan Du[2*] and Wei Jiang[1]
[1]Antai College of Economics and Management, Shanghai Jiao Tong University, Shanghai, China
[2]Smart Manufacturing Thrust, Systems Hub, The Hong Kong University of Science and Technology, Guangzhou, China

**Abstract**

Design of process control scheme is critical for quality assurance to reduce variations in manufacturing systems. Taking semiconductor manufacturing as an example, extensive literature focuses on control optimization based on certain process models (usually linear models), which are obtained by experiments before a manufacturing process starts. However, in real applications, pre-defined models may not be accurate, especially for a complex manufacturing system. To tackle model inaccuracy, we propose a model-free reinforcement learning (MFRL) approach to conduct experiments and optimize control simultaneously according to real-time data. Specifically, we design a novel MFRL control scheme by updating the distribution of disturbances using Bayesian inference to reduce their large variations during manufacturing processes. As a result, the proposed MFRL controller is demonstrated to perform well in a nonlinear chemical mechanical planarization (CMP) process when the process model is unknown. Theoretical properties are also guaranteed when disturbances are additive. The numerical studies also demonstrate the effectiveness and efficiency of our methodology.

## 1. INTRODUCTION

### 1.1 *Background and motivations*

Process control is critical to keep the stability of manufacturing processes and guarantee the quality of final products, especially when a manufacturing process is complex. For example, in a semiconductor manufacturing process, two types of factors influence the stability of the manufacturing system. First, internal factors from manufacturing equipment and environments, mainly refer to process dynamics and disturbances during the manufacturing process (Tseng and Chen, 2017). Second, external factors refer to control recipes designed by the manufacturer, which aim to compensate for disturbances and adjust the system output to its desired target.

Traditional run-to-run (R2R) control schemes can be divided into two phases. In Phase I, a process model is specified to describe the relationship between control input and process output through domain knowledge, design of experiments (DOE), or response surface methodology (RSM), followed by control recipe optimizations in Phase II (Tseng et al., 2019). A detailed literature review is provided in Section 1.2. However, in practical applications, when manufacturing processes are too complex to be described by specific models accurately, traditional R2R controllers may encounter significant challenges in accurate quality control. For example, chemical mechanical planarization (CMP) process is one of the most important steps in semiconductor manufacturing to remove excess materials from the surface of silicon wafers. In literature, CMP processes are often controlled with explicit assumptions of process models (Castillo and Yeh, 1998). However, such models cannot fully capture the relationship between system outputs, control recipes, and disturbances, thereby leading to unavoidable model errors, which affect the accuracy of control optimization.

To tackle model inaccuracy in complex manufacturing processes, model-free reinforcement learning (MFRL) approaches (Recht, 2019) have been developed to learn manufacturing environments from real-time experimental data and directly search optimal control recipes without process model assumptions. Therefore, MFRL provides unprecedented opportunities for control optimization, especially in complex manufacturing processes. However, current MFRL approaches need to be improved as disturbances are hidden unstable factors that affect system outputs significantly (Nian et al., 2020). Take CMP process as an example, Figure 1 illustrates the system outputs based on the MFRL controller in Recht (2019) (defined as a basic MFRL controller). In the basic MFRL controller, the effects of disturbances are ignored and control recipes are directly optimized based on system outputs. As shown in Figure 1, compared with the case without control, the basic MFRL controller can roughly keep the system output close to the target level. However, the controlled process still experiences significant deviations during some periods, which leads to invalid control. Therefore, it is highly desired to design a new control methodology to improve the basic MFRL controller by updating real-time distributions of disturbances to reduce the variations.

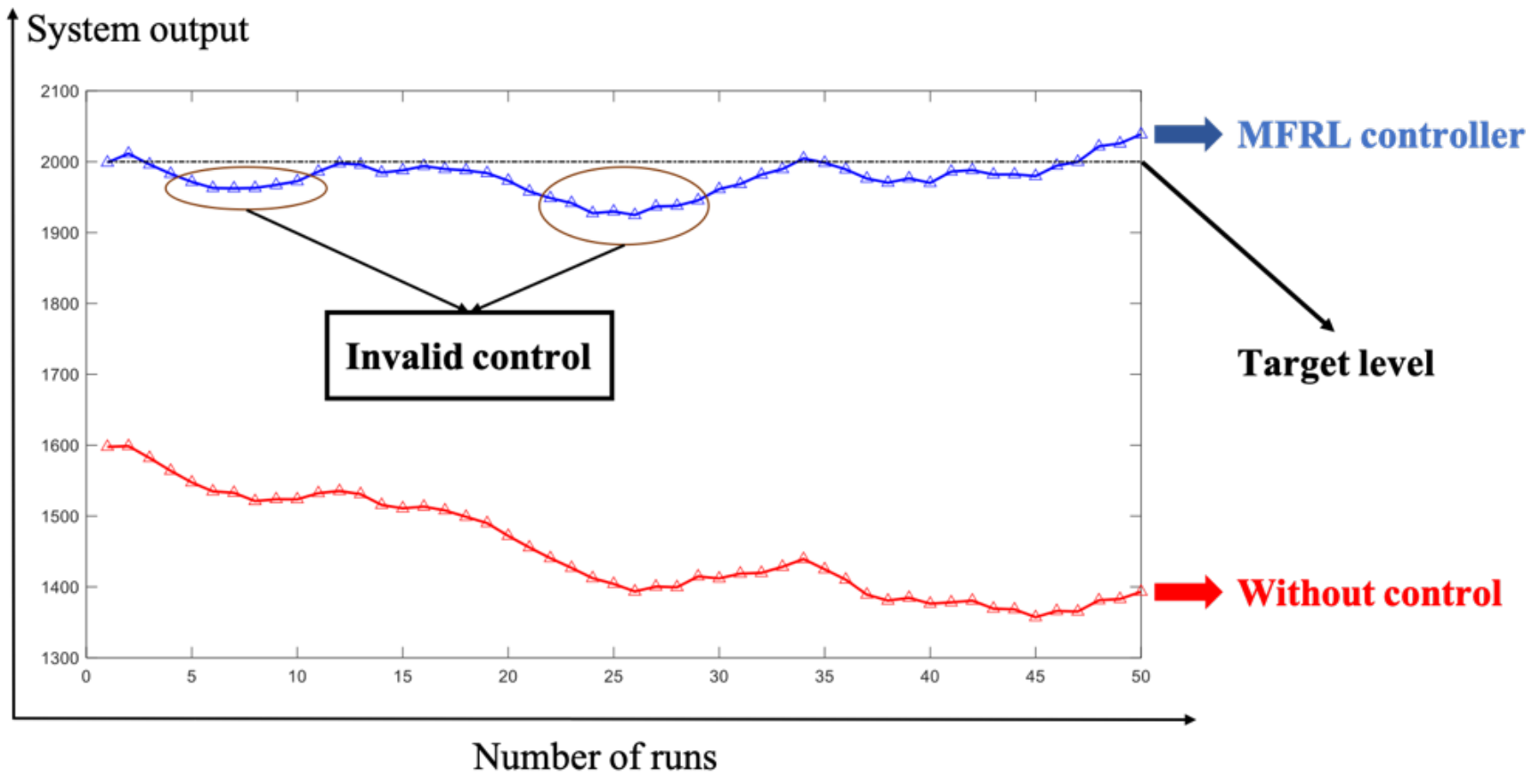

Figure 1. An example of basic MFRL controller in a CMP process

## 1.2 *Literature review*

In this subsection, we review different process control methods for complex manufacturing systems, especially for semiconductor manufacturing. Since the control mechanism or process model is important for controller design (Bastiaan, 1997), we classify the literature into two main categories based on whether the process model is available/predefined or not: (1) model-based controllers and (2) data-driven or model-free controllers.

Both linear and nonlinear process models have been considered in existing process control methodologies. Extensive pioneer works considered linear process models with disturbances that follow different stochastic time series. For example, Ingolfsson and Sachs (1993) analyzed the stability and sensitivity of the exponentially weighted moving average (EWMA) controller in compensating for the integrated moving average (IMA) disturbance process. Ning et al. (1996) formulated the process model as a linear transfer function with time-dependent drifts and developed a time-based EWMA controller. Tsung and Shi (1999) designed a proportional-integral-derivative (PID) controller for linear process models with autoregressive moving average (ARMA) disturbances and integrated the PID-based control scheme with statistical process control. Chen and Guo (2001) proposed an age-based double EWMA controller, which performs better than the EWMA controller in dealing with time-dependent drifts. Tseng et al. (2003) designed a new controller to improve the traditional EWMA controller by optimizing its discount factor and defined it as the variable-EWMA (VEWMA) controller, which has great performance in linear process models with ARIMA disturbance. Tseng et al. (2007) showed that the VEWMA controller has better performance than double EWMA numerically. He et al. (2009)

proposed a new controller named general harmonic rule (GHR) and theoretically proved its performance for a wide range of stochastic disturbances.

Besides linear process models, nonlinear process models are also widely studied. Hankinson et al. (1997) introduced a polynomial function to approximate a process model in deep reactive ion etching. Del Castillo and Yeh (1998) reviewed different polynomial process models for approximation of the CMP process and proposed adaptive R2R controllers according to these polynomial models. Kazemzadeh et al. (2008) extended the EWMA and VEWMA controllers in quadratic process models. In addition to polynomial models, more complicated nonlinear process models are introduced by differential equations. For example, Bibian and Jin (2000) considered a digital control problem in a second-order system and proposed two practical control schemes to deal with the time delay. Chen et al. (2012) focused on the deterministic as well as stochastic process models with measurement delay and proposed a new controller that integrates deterministic and stochastic components with applications in chemical vapor deposition (CVD) processes. In summary, model-based controllers depend crucially on explicit process formulations and are suitable for cases where the focused process models are well-validated.

When an explicit process model is not available, data-driven or model-free controllers are directly designed based on historical or offline data. For example, neural networks (NN) are widely used to approximate the unknown process model according to control recipes and system outputs. Park et al. (2005) approximated the real process model by an NN and designed an NN-based controller to reduce overlay misalignment errors significantly in semiconductor manufacturing processes. Wang and Chou (2005) proposed a neural-Taguchi-based control strategy to reach the desired material removal rate through an NN-simulated CMP process. Chang et al. (2006) developed a virtual metrology system using different NNs to describe the process model and optimized the control recipes accordingly. Liu et al. (2018) summarized NN-based controllers in their review paper and emphasized the related practical issues such as nonstationary control results and poor interpretations. Therefore, when controlling dynamic manufacturing systems characterized by unstable disturbances, existing NN-based approaches also encounter challenges in accurately approximating the manufacturing process.

Compared with NN-based control methods, reinforcement learning (RL) is another efficient data-driven control method to learn system dynamics and optimize control recipes by interacting with real-time system states. Given the definition of system state, control policy, and cost or reward function, RL can optimize control recipes based on real-time system states (Wang et al., 2018). For example, Recht (2019) introduced two basic policy-based algorithms for MFRL methods, policy gradient and pure random search (PRS). The policy gradient method optimizes control strategies based on the distribution of system outputs (Li et al., 2023), while the PRS method is more general and directly optimizes control strategies by stochastic gradient descent. However, as pointed out by Nian et al. (2020), these MFRL controllers cannot be directly applied in complex manufacturing systems due to large variations caused by unknown process models and unstable disturbances. Therefore, Khamaru et al. (2021) explored an effective variance reduction method based on an instance-dependent function in Q-learning.

In summary, the above data-driven methods share a common limitation that variations are relatively large. As process models are unknown, hidden unstable disturbances are hard to be recognized, thereby bringing difficulties to optimize control recipes compensating for them. To tackle the challenges, in this article, we design a new process control scheme to improve the basic MFRL controller (e.g., PRS-based MFRL controller) by updating the distribution of disturbances through Bayesian inference. We define it as a model-free reinforcement learning controller with Bayesian inference (MFRL-BI).

As disturbances can be reflected by system outputs, we use Bayesian inference to update the real-time distribution and integrate it into current MFRL control schemes. Figure 2 illustrates the difference between the control schemes of existing R2R and the proposed MFRL-BI controllers in terms of process assumptions and control optimization. Following the design steps of process control scheme in Figure 2 (Del Castillo and Hurwitz, 1997), we divide the MFRL-BI controller into two phases: the optimization phase for controller learning (Phase I) and the application phase in online manufacturing (Phase II). In Phase I, we design experiments by virtual metrology (VM) to provide extensive data (Chang et al., 2006; Kang et al., 2009) for searching control recipes using MFRL algorithms. Considering the fact that disturbance can be inferred by system outputs, we update its distribution through Bayesian inference using real-time outputs. Finally, the input control recipes, system outputs, and disturbance inference data are collected and used for online control in Phase II.

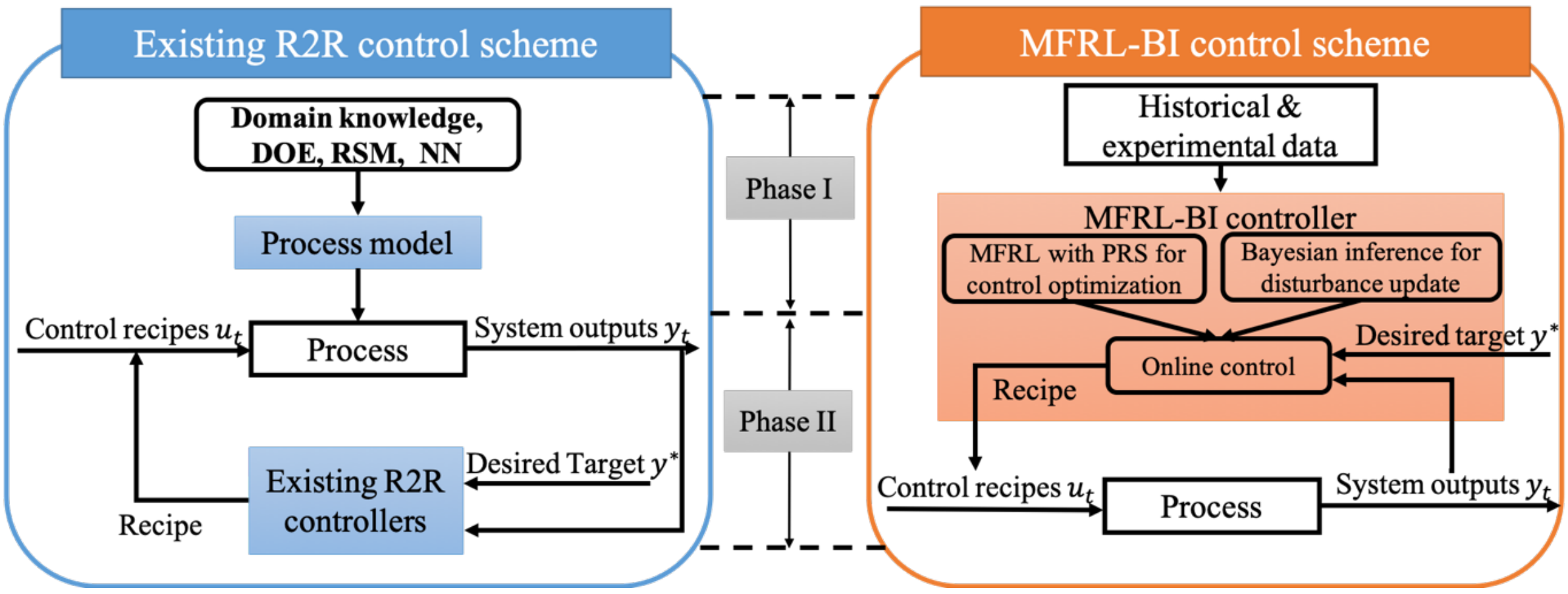

Figure 2. Difference between existing R2R and MFRL-BI control schemes

The main contributions of our work are summarized as follows: (1) a new model-free control scheme called MFRL-BI is proposed for efficient variation reduction by updating disturbance processes through Bayesian inference. (2) The corresponding algorithms of the MFRL-BI controller that combine Bayesian inference with the current PRS-based MFRL methodology are presented. (3) The proposed MFRL-BI controller is theoretically shown to guarantee optimality asymptotically.

The remainder of this paper is organized as follows. Section 2 introduces the basic MFRL methodology in an R2R control scheme. Section 3 provides the design procedure of the MFRL-BI control scheme and interprets the related theoretical principles in Phases I and II. Section 4 demonstrates the performance of our method numerically and compares it with the DOE-based automatic process controller (APC) with the application in a nonlinear CMP process control. Finally, Section 5 concludes the paper with remarks on future research directions.

## 2. BASIC MFRL CONTROLLER

In this section, we first present formulations of the process control problem in Section 2.1, and then discuss the methodology and corresponding algorithms of the basic MFRL in Section 2.2.

### 2.1 *Process control formulation*

We consider a multiple input-multiple output (MIMO) R2R process control problem that aims to reduce variations in a manufacturing system. At run $t \in \{1,2, \dots T\}$, a control recipe $\boldsymbol{u}_t \in \mathbb{R}^{m\times 1}$ is optimized to keep the system output $\boldsymbol{y}_t \in \mathbb{R}^{n\times 1}$ close to its target level $\boldsymbol{y}^* \in \mathbb{R}^{n\times 1}$, where $T$ is the total number of runs. $m$ and $n$ are the dimensions of input control recipes and system outputs, respectively. The squared errors of process outputs are used to measure the control cost (Wang and Han, 2013). Furthermore, as control actions also bring extra costs in the manufacturing process, the cost function at run $t$ is:

$$C_t(\boldsymbol{y}_t, \boldsymbol{u}_t) = (\boldsymbol{y}_t - \boldsymbol{y}^*)^{\boldsymbol{T}} \boldsymbol{Q} (\boldsymbol{y}_t - \boldsymbol{y}^*) + \boldsymbol{u}_t^{\boldsymbol{T}} \boldsymbol{R} \boldsymbol{u}_t, \tag{1}$$

where $\boldsymbol{Q}$ and $\boldsymbol{R}$ are positive definite weighted matrices. According to Del Castillo and Hurwitz (1997), the system output $\boldsymbol{y}_t$ is affected by the control recipes $\boldsymbol{u}_t$ as well as disturbances in manufacturing environments. Therefore, we define the underlying truth of the unknown process model as $\boldsymbol{y}_t = h(\boldsymbol{u}_t, \boldsymbol{d}_t)$, where $\boldsymbol{d}_t \in \mathbb{R}^{n\times 1}$ is the disturbance at run $t$. Combining with the cost function in Equation (1), we have the process control problem in $T$ runs as:

$$\min_{\{\boldsymbol{u}_1, \boldsymbol{u}_2, \ldots, \boldsymbol{u}_T\}} E_{\{\boldsymbol{d}_1, \boldsymbol{d}_2, \ldots, \boldsymbol{d}_T\}} \left[ \sum_{t=1}^{T} \left( (\boldsymbol{y}_t - \boldsymbol{y}^*)^{\boldsymbol{T}} \boldsymbol{Q} (\boldsymbol{y}_t - \boldsymbol{y}^*) + \boldsymbol{u}_t^{\boldsymbol{T}} \boldsymbol{R} \boldsymbol{u}_t \right) \right]$$

$$\text{s.t. } \boldsymbol{y}_t = h(\boldsymbol{u}_t, \boldsymbol{d}_t). \tag{2}$$

Note that the process model $h(\boldsymbol{u}_t, \boldsymbol{d}_t)$ is general and not specified.

In semiconductor manufacturing, it is widely recognized that process disturbances come from manufacturing systems or environments, both of which are independent of control recipes. Meanwhile, the effects of control recipes and disturbances are additive in a process model (Box and Kramer, 1992; Zhong et al, 2010; Wang and Han, 2013). Therefore, we have Assumption 2.1.

**Assumption 2.1**: *The manufacturing process outputs can be separated into two additive parts related to control recipes and disturbances respectively, i.e.,*

$$\boldsymbol{y}_t = h(\boldsymbol{u}_t, \boldsymbol{d}_t) = g(\boldsymbol{u}_t) + \boldsymbol{d}_t. \tag{3}$$

*where* $g(\boldsymbol{u}_t)$ *and* $\boldsymbol{d}_t$ *are assumed to be independent.*

In semiconductor manufacturing systems, disturbance processes exhibit general autocorrelations due to manufacturing environments such as aging effects (Del Castillo and Hurwitz, 1997). Therefore, in a manufacturing cycle from runs 1 to $T$, the disturbance $\boldsymbol{d}_t$ can be inferred from its historical trajectory $\boldsymbol{D}_{t-1} = [\boldsymbol{d}_1, \boldsymbol{d}_2, \ldots \boldsymbol{d}_{t-1}]$. We define the conditional probability density function of the disturbance at run $t$ as $p(\boldsymbol{d}_t | \boldsymbol{D}_{t-1})$ with mean vector $\boldsymbol{\mu}_t$ and covariance matrix $\boldsymbol{\Sigma}_t$.

For control recipes to compensate for the disturbances, as shown in Equation (3), their effects on the system output are modeled by a function $g(\cdot)$, which is often assumed as a linear function in literature (Chen and Guo, 2001; Tseng et al., 2003; 2007). Considering the potential inaccuracy, we relax formulation assumptions of $g(\cdot)$ in our model. Although the effects of control recipes and disturbances on the system output are separated according to Assumption 2.1, there still exists a significant challenge in quantifying the effects of control recipes and disturbances as $g(\cdot)$ is unknown and $\boldsymbol{d}_t$ cannot be observed directly.

## 2.2 *Methodology of basic MFRL with PRS*

In the control methodology of a basic MFRL controller, the expectation of control cost over disturbances $\boldsymbol{d}_t$ is minimized by optimizing control recipe $\boldsymbol{u}_t$. Due to the unknown process model $g(\cdot)$, the cost function is also an unknown function over $\boldsymbol{u}_t$. According to Recht (2019), the objective function in Equation (2) can be reformulated as $J(\boldsymbol{u}) = \mathbf{E}_{\{\boldsymbol{d}_1,\boldsymbol{d}_2,\ldots,\boldsymbol{d}_T\}}[\sum_{t=1}^{T} C_t(\boldsymbol{y}_t(\boldsymbol{u}_t, \boldsymbol{d}_t), \boldsymbol{u}_t)]$, where $\boldsymbol{u} = [\boldsymbol{u}_1, \ldots, \boldsymbol{u}_t, \ldots \boldsymbol{u}_T]$. Before optimizing the function $J(\boldsymbol{u})$, suppose the following assumption holds.

**Assumption 2.2**: *The function* $J(\boldsymbol{u}) = \mathbf{E}_{\{\boldsymbol{d}_1,\boldsymbol{d}_2,\ldots,\boldsymbol{d}_T\}}[\sum_{t=1}^{T} C_t(\boldsymbol{y}_t(\boldsymbol{u}_t, \boldsymbol{d}_t), \boldsymbol{u}_t)]$ *achieves a minimum at an unknown point* $\boldsymbol{u}^*$.

To minimize $J(\boldsymbol{u})$, the basic MFRL controller in Recht (2019) uses a PRS-based method to optimize the control recipes by stochastic gradient descent (SGD). If Assumptions 2.1 and 2.2 hold, the optimization problem in Equation (2) can be solved via the SGD algorithm as follows.

**SGD Algorithm**: *There are two steps in the SGD algorithm for the basic MFRL controller. First, the gradient of* $J(\boldsymbol{u})$ *is approximated by a finite difference along the direction* $\boldsymbol{\epsilon}$, *where* $\boldsymbol{\epsilon} \in \mathbb{R}^{m\times T}$ *is a random vector consisting of 0 or 1. Then, we can write the gradient of* $J(\boldsymbol{u})$ *as:*

$$\nabla_{\boldsymbol{u}} J(\boldsymbol{u}) = \frac{J(\boldsymbol{u}+\iota\boldsymbol{\epsilon}) - J(\boldsymbol{u}-\iota\boldsymbol{\epsilon})}{2\iota}\boldsymbol{\epsilon}, \tag{4}$$

*where* $\iota \to 0$ *and* $\boldsymbol{u} \mp \iota\boldsymbol{\epsilon}$ *denote the neighborhood of the control strategy* $\boldsymbol{u}$. *Second, the control recipe moves along the gradient descent direction with step size* $\alpha$. *If* $\boldsymbol{u}^{[k]}$ *is used to denote the value of control recipes in the* $k$*th iteration, we have*

$$\boldsymbol{u}^{[k+1]} = \boldsymbol{u}^{[k]} - \alpha\nabla_{\boldsymbol{u}} J(\boldsymbol{u}^{[k]}). \tag{5}$$

*These two steps are executed alternately until* $\boldsymbol{u}$ *converges (i.e., the difference between successive iterated values of* $\boldsymbol{u}^{[k+1]}$ *and* $\boldsymbol{u}^{[k]}$ *is smaller than a pre-defined threshold* $\eta$*).*

Following the SDG algorithm, Algorithm 1 presents the aforementioned control search procedure to minimize the unknown function $J(\cdot)$.

---

**Algorithm 1. MFRL with PRS Algorithm**

---

Function: MFRL_PRS(·)

Input: hyper-parameters $\boldsymbol{\epsilon}$, $\iota$, $\alpha$, $\eta$

Initialize: $k = 0$, control recipe $\boldsymbol{u}^{[0]}$

**Repeat**:

Execute two initial control strategies $\boldsymbol{u}^{[k]} + \iota\boldsymbol{\epsilon}$ and $\boldsymbol{u}^{[k]} - \iota\boldsymbol{\epsilon}$

$\nabla_{\boldsymbol{u}} J\left(\boldsymbol{u}^{[k]}\right) = \frac{J(\boldsymbol{u}^{[k]}+\iota\boldsymbol{\epsilon}) - J(\boldsymbol{u}^{[k]}-\iota\boldsymbol{\epsilon})}{2\iota}\boldsymbol{\epsilon}$

$\boldsymbol{u}^{[k+1]} = \boldsymbol{u}^{[k]} - \alpha\nabla_{\boldsymbol{u}} J\left(\boldsymbol{u}^{[k]}\right)$

$k \leftarrow k + 1$

**Until** $\left\|\boldsymbol{u}^{[k]} - \boldsymbol{u}^{[k-1]}\right\| < \eta$

$\hat{\boldsymbol{u}} = \boldsymbol{u}^{[k]}$

Output: $\hat{\boldsymbol{u}}$

---

According to the asymptotic analysis of SGD algorithm in Kiefer and Wolfowitz (1952), if disturbances satisfy the condition $\mathbf{E}(\boldsymbol{d}_t) = \mathbf{0}$, the control recipe searched in Algorithm 1 will converge to the optimal value. However, in practice, the disturbance process is not stable, its fluctuations and drifts are inevitable and may even increase as time goes by. For example, in CMP process in Figure 1, the basic MFRL controller encounters large variations, as it focuses on minimizing the expected control cost $J(\boldsymbol{u})$ but ignores the variations and drifts of disturbance $\boldsymbol{d}_t$. To overcome this limitation, we propose the MFRL-BI controller to further reduce the variations of system outputs by dynamically updating the distribution of disturbances in Section 3.

# 3. THE MFRL-BI CONTROLLER

In this section, the MFRL-BI controller is proposed to improve the performance of basic MFRL by updating the distribution of disturbance via Bayesian inference. Following Figure 2, we introduce methodologies of the proposed MFRL-BI controller in two phases in Sections 3.1 and 3.2 respectively. As shown in Figure 3, in Phase I, control recipes are searched in the inner loop using the MFRL algorithm with PRS. After taking the convergent control recipe, the distribution of disturbance is updated in the outer loop. Meanwhile, the control recipes, system outputs, and estimated disturbances are collected, which are used for online control optimization in Phase II.

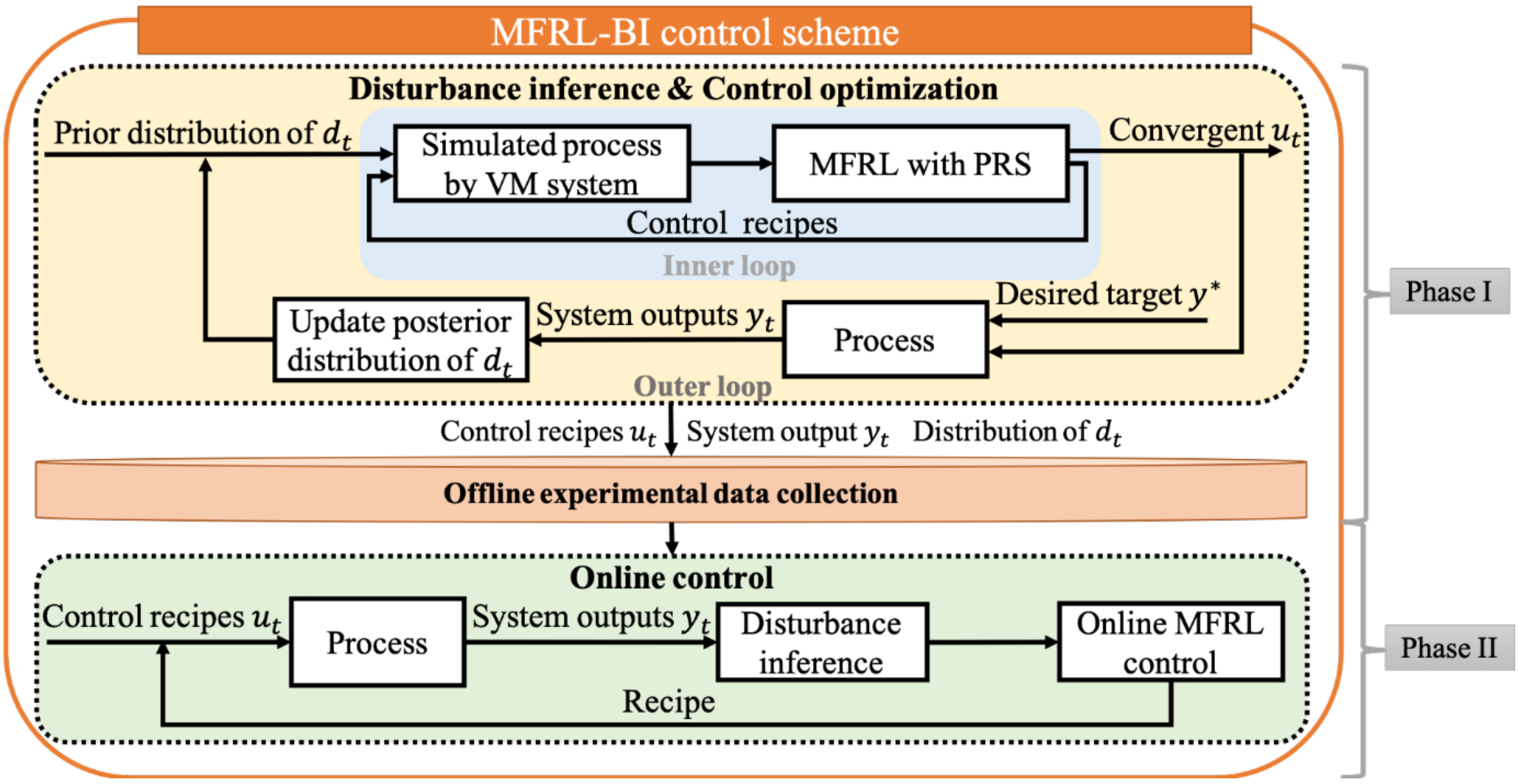

Figure 3. The methodology of the MFRL-BI controller

As introduced in Section 2.2, disturbances are unobservable, we define the prior distribution of $\boldsymbol{d}_t$ condition on its trajectory as

$$\boldsymbol{d}_t|\boldsymbol{D}_{t-1} \sim p(\boldsymbol{\mu}_t, \boldsymbol{\Sigma}_t), \tag{6}$$

where $p(\cdot)$ is the probability distribution function. The observations of system output $\boldsymbol{y}_t$ can reflect the disturbance process and be used to update the posterior distribution of $\boldsymbol{d}_t$. However, $\boldsymbol{y}_t$ is also affected by the control recipe $\boldsymbol{u}_t$, which brings challenges for disturbance inference. Therefore, in Figure 3, we separate the effects of $\boldsymbol{d}_t$ and $\boldsymbol{u}_t$, and make inference of $\boldsymbol{d}_t$ in the outer loop and optimization of $\boldsymbol{u}_t$ in the inner loop.

Specifically, to separate the effects of $\boldsymbol{d}_t$ and $\boldsymbol{u}_t$, we reformulate the process model in Equation (3) as $\boldsymbol{y}_t = g(\boldsymbol{u}_t) + \boldsymbol{d}_t = g(\boldsymbol{u}_t) + \boldsymbol{\mu}_t + \boldsymbol{\delta}_t$, where $\boldsymbol{\mu}_t$ is the mean vector of $\boldsymbol{d}_t$ and $\boldsymbol{\delta}_t = \boldsymbol{d}_t - \boldsymbol{\mu}_t$ is a random vector with $E(\boldsymbol{\delta}_t) = \mathbf{0}$. Since the process model $g(\boldsymbol{u}_t)$ is unknown, the variability of searched control recipe via Algorithm 1 using PRS is unavoidable, especially when the number of iterations is limited and the step size is fixed (Kiefer and Wolfowitz, 1952). We use $\boldsymbol{v}_t = \widehat{\boldsymbol{u}}_t - \boldsymbol{u}_t^*$ to denote this variability, where $\widehat{\boldsymbol{u}}_t$ is control recipe searched by PRS and $\boldsymbol{u}_t^*$ is the underlying optimal control recipe. In summary, we reformulate the optimization problem in Equation (2) as follows at each run $t$:

$$\min_{\boldsymbol{u}_t} \mathbf{E}_{\boldsymbol{\delta}_t,\boldsymbol{v}_t}[C_t(\boldsymbol{y}_t, \boldsymbol{u}_t)]$$
$$\text{s.t. } \boldsymbol{y}_t = g(\boldsymbol{u}_t) + \boldsymbol{\mu}_t + \boldsymbol{\delta}_t. \tag{7}$$

By incorporating the constraints into the objective function, we have:

$$\mathbf{E}_{\boldsymbol{\delta}_t,\boldsymbol{v}_t}[C_t(\boldsymbol{y}_t, \boldsymbol{u}_t)] = tr(\boldsymbol{Q}\boldsymbol{\Sigma}_t) + \mathbf{E}_{\boldsymbol{v}_t}[(g(\boldsymbol{u}_t) + \boldsymbol{\mu}_t - \boldsymbol{y}^*)^T \boldsymbol{Q}(g(\boldsymbol{u}_t) + \boldsymbol{\mu}_t - \boldsymbol{y}^*)] + \boldsymbol{u}_t^T \boldsymbol{R} \boldsymbol{u}_t. \tag{8}$$

Detailed derivations are presented in Appendix A. For convenience, we define the function $M(\boldsymbol{u}_t|\boldsymbol{\mu}_t)$ given the distribution of disturbances as:

$$M(\boldsymbol{u}_t|\boldsymbol{\mu}_t) := \mathbf{E}_{\boldsymbol{v}_t}[(g(\boldsymbol{u}_t) + \boldsymbol{\mu}_t - \boldsymbol{y}^*)^T \boldsymbol{Q}(g(\boldsymbol{u}_t) + \boldsymbol{\mu}_t - \boldsymbol{y}^*)] + \boldsymbol{u}_t^T \boldsymbol{R} \boldsymbol{u}_t, \tag{9}$$

Then the total cost can be divided into two parts: $M(\boldsymbol{u}_t|\boldsymbol{\mu}_t)$ and $tr(\boldsymbol{Q}\boldsymbol{\Sigma}_t)$. This separation allows us to optimize $M(\boldsymbol{u}_t|\boldsymbol{\mu}_t)$ by MFRL algorithm with PRS and update the value of $tr(\boldsymbol{Q}\boldsymbol{\Sigma}_t)$ and $\boldsymbol{\mu}_t$ by Bayesian inference. The methodology and corresponding algorithms of control optimization and disturbance inference in Phase I will be elaborated in Section 3.1.

## 3.1 *Control optimization in Phase I*

To separate the effects of $\boldsymbol{u}_t$ and $\boldsymbol{d}_t$, we divide the control process at each run into two steps: (i) at the beginning of run $t$, given the prior distribution of $\boldsymbol{d}_t$, control recipe $\boldsymbol{u}_t$ is searched to minimize the control cost $M(\boldsymbol{u}_t|\boldsymbol{\mu}_t)$; (ii) the posterior distribution of $\boldsymbol{d}_t$ is updated when the system output $\boldsymbol{y}_t$ is observed and the prior distribution of $\boldsymbol{d}_{t+1}$ is inferred according to the posterior distribution of $\boldsymbol{d}_t$. These two steps correspond to the inner and outer loops in Figure 3, respectively, and are presented as follows.

## *A. Inner loop: search for control recipes*

In this part, we design an experiment searching for control recipes to minimize the expected control cost $M(\boldsymbol{u}_t|\boldsymbol{\mu}_t)$. According to its definition in Equation (9), we can separate $M(\boldsymbol{u}_t|\boldsymbol{\mu}_t)$ as:

$$M(\boldsymbol{u}_t|\boldsymbol{\mu}_t) \coloneqq H(\boldsymbol{u}_t|\boldsymbol{\mu}_t) + \boldsymbol{u}_t^T \boldsymbol{R} \boldsymbol{u}_t, \tag{10}$$

where $H(\boldsymbol{u}_t|\boldsymbol{\mu}_t) = [(g(\boldsymbol{u}_t) + \boldsymbol{\mu}_t - \boldsymbol{y}^*)^T \boldsymbol{Q}(g(\boldsymbol{u}_t) + \boldsymbol{\mu}_t - \boldsymbol{y}^*)]$. As $\boldsymbol{u}_t^T \boldsymbol{R} \boldsymbol{u}_t$ is a deterministic convex function of $\boldsymbol{u}_t$, it is necessary to search the gradient of $H(\cdot)$, and we have $\nabla_{\boldsymbol{u}_t} M(\boldsymbol{u}_t|\boldsymbol{\mu}_t) = \nabla_{\boldsymbol{u}_t} H(\boldsymbol{u}_t|\boldsymbol{\mu}_t) + 2\boldsymbol{R}\boldsymbol{u}_t$. Before searching for $\boldsymbol{u}_t$, we suppose that $H(\cdot)$ also satisfies Assumption 2.2, i.e., $H(\cdot)$ is an unknown function that has a minimum at an unknown point $\tilde{\boldsymbol{u}}_t$ ($\tilde{\boldsymbol{u}}_t = arg \min_{\boldsymbol{u}_t} H(\boldsymbol{u}_t|\boldsymbol{\mu}_t)$). Then, similar to the basic MFRL controller, we implement Algorithm 1 to optimize the unknown function $M(\cdot)$ using PRS. Particularly, to further guarantee the stability of control recipes and reduce the variability of $\boldsymbol{v}_t$, after the convergence of $\boldsymbol{u}_t$ based on Algorithm 1, we execute another $N$ iterations of control recipes, which are denoted as $\hat{\boldsymbol{u}}_t(1)$ to $\hat{\boldsymbol{u}}_t(N)$. The final recipe is chosen as the mean of control recipes after convergence (i.e., $\bar{\boldsymbol{u}}_t = \frac{1}{N}\sum_{i=1}^{N} \hat{\boldsymbol{u}}_t(i)$). Algorithm 2 presents the details of the control optimization in the MFRL-BI controller.

**Algorithm 2. Control optimization given disturbance distribution**

Function: Control_Search

Input: parameter $\boldsymbol{\mu}_t$, hyper-parameters $\boldsymbol{\epsilon} \in \mathbb{R}^{m\times 1}$, $\alpha$, $N$, $\iota$

Output: $\bar{\boldsymbol{u}}_t$

Initialize: control recipe $\boldsymbol{u}_t^{[0]}$

Calculate $\hat{\boldsymbol{u}}_t(1)$ using Algorithm 1 based on function $M(\cdot\,|\boldsymbol{\mu}_t)$

**For** $i = 1$ to $N-1$ do

Execute control strategies $\hat{\boldsymbol{u}}_t(i) + \iota\boldsymbol{\epsilon}$ and $\hat{\boldsymbol{u}}_t(i) - \iota\boldsymbol{\epsilon}$

$\nabla_{\boldsymbol{u}_t} M(\hat{\boldsymbol{u}}_t(i)|\boldsymbol{\mu}_t) = \frac{H(\hat{\boldsymbol{u}}_t(i)+\iota\boldsymbol{\epsilon}|\boldsymbol{\mu}_t)+H(\hat{\boldsymbol{u}}_t(i)-\iota\boldsymbol{\epsilon}|\boldsymbol{\mu}_t)}{2\iota}\boldsymbol{\epsilon} + 2\boldsymbol{R}\hat{\boldsymbol{u}}_t(i)$

$\hat{\boldsymbol{u}}_t(i+1) = \hat{\boldsymbol{u}}_t(i) - \alpha\nabla_{\boldsymbol{u}_t} M(\hat{\boldsymbol{u}}_t(i)|\boldsymbol{\mu}_t)$

**End for**

$\bar{\boldsymbol{u}}_t = \frac{1}{N}\sum_{i=1}^{N} \hat{\boldsymbol{u}}_t(i)$

Algorithm 2 has two procedures: first, control recipes are searched to minimize the cost function $M(\cdot)$ given the distribution of disturbances. Second, after the convergence of control recipes, we use another $N$ samples to reduce the variations of control resulting from stochastic gradient approximation for the unknown function $H(\cdot)$. To further examine the properties of searched control recipes in Algorithm 2, we make two assumptions about function $H(\cdot)$ as in Mandt et al. (2017).

**Assumption 3.1:** *The stochastic gradient in Algorithm 2 can be expressed as the underlying truth gradient value plus a random gradient noise. The noise can be approximated as Gaussian, whose variance is independent of control recipes. i.e.,* $\nabla_{\boldsymbol{u}_t} H(\boldsymbol{u}_t|\boldsymbol{\mu}_t) \approx \nabla_{\boldsymbol{u}_t} H^*(\boldsymbol{u}_t|\boldsymbol{\mu}_t) + \boldsymbol{\varepsilon}$ *and* $\nabla_{\boldsymbol{u}_t} M(\boldsymbol{u}_t|\boldsymbol{\mu}_t) \approx \nabla_{\boldsymbol{u}_t} M^*(\boldsymbol{u}_t|\boldsymbol{\mu}_t) + \boldsymbol{\varepsilon}$*, where* $\nabla_{\boldsymbol{u}_t} H^*(\boldsymbol{u}_t|\boldsymbol{\mu}_t)$ *and* $\nabla_{\boldsymbol{u}_t} M^*(\boldsymbol{u}_t|\boldsymbol{\mu}_t)$ *denote the underlying truth gradients of functions* $H(\cdot)$ *and* $M(\cdot)$*, respectively. It is obvious that* $\nabla_{\boldsymbol{u}_t} M^*(\boldsymbol{u}_t|\boldsymbol{\mu}_t) = \nabla_{\boldsymbol{u}_t} H^*(\boldsymbol{u}_t|\boldsymbol{\mu}_t) + 2\boldsymbol{R}\boldsymbol{u}_t$ *according to their definition.* $\boldsymbol{\varepsilon}$ *follows a multi-normal distribution with zero mean vector and covariance matrix* $\boldsymbol{\Sigma}_{\boldsymbol{\varepsilon}}$.

**Assumption 3.2:** *The finite-difference equation of control iterations can be approximated by the stochastic differential equation. Specifically, the difference equation between two successive control iterations searched by Algorithm 2 (*$\Delta\boldsymbol{u}_t = -\alpha\nabla_{\boldsymbol{u}_t} M(\boldsymbol{u}_t|\boldsymbol{\mu}_t)$*) can be approximated by* $\mathrm{d}\boldsymbol{u}_t = -\alpha\nabla_{\boldsymbol{u}_t} M(\boldsymbol{u}_t|\boldsymbol{\mu}_t)\mathrm{d}t$*. Combined with Assumption 3.1, we have* $\mathrm{d}\boldsymbol{u}_t = -\alpha\nabla_{\boldsymbol{u}_t} M^*(\boldsymbol{u}_t|\boldsymbol{\mu}_t)\mathrm{d}t + \alpha\boldsymbol{B}\mathrm{d}W_t$*, where* $\boldsymbol{B}^{\boldsymbol{T}}\boldsymbol{B} = \boldsymbol{\Sigma}_{\boldsymbol{\varepsilon}}$ *and* $W_t$ *is a standard Wiener process.*

According to Assumptions 3.1 and 3.2 on the unknown functions $H(\cdot)$, Theorem 1 shows the theoretical property of the searched control recipes in Algorithm 2.

**Theorem 1**: *The searched control recipe using Algorithm 2 is asymptotically optimal.*

The proof is provided in Appendix B.

Theorem 1 guarantees the asymptotic optimality of Algorithm 2 when process models are unknown for complex manufacturing processes in general. Specifically, if the function $H(\boldsymbol{u}_t|\boldsymbol{\mu}_t)$ can also be approximated by its second-order Taylor expansion, more theoretical properties are obtained related to the closed-form solution (Proposition 1), the stochastic searching process (Theorem 2), and the stationary distribution (Theorem 3) of the control recipes.

**Proposition 1**: *If function* $H(\boldsymbol{u}_t|\boldsymbol{\mu}_t)$ *has a minimum at an unknown point* $\widetilde{\boldsymbol{u}}_t$ *,i.e.,* $\widetilde{\boldsymbol{u}}_t := arg \min_{\boldsymbol{u}_t} H(\boldsymbol{u}_t|\boldsymbol{\mu}_t)$ *, the optimal control recipe to minimize the cost* $C_t$ *is* $\boldsymbol{u}_t^* = (\boldsymbol{G}^T\boldsymbol{Q}\boldsymbol{G} + \boldsymbol{R})^{-1}\boldsymbol{G}^T\boldsymbol{Q}\boldsymbol{G}\widetilde{\boldsymbol{u}}_t$*, where* $\boldsymbol{G} = \begin{bmatrix} \frac{\partial g_1}{\partial \tilde{u}_1} & \cdots & \frac{\partial g_1}{\partial \tilde{u}_m} \\ \vdots & \ddots & \vdots \\ \frac{\partial g_n}{\partial \tilde{u}_1} & \cdots & \frac{\partial g_n}{\partial \tilde{u}_m} \end{bmatrix}_{n\times m}$ *is the gradient matrix of function* $g(\cdot)$*.*

The proof is provided in Appendix C.

**Theorem 2**: *The control search process for* $\boldsymbol{u}_t^*$ *in Algorithm 2 can be approximated by an Ornstein-Uhlenbeck process, i.e.,* $d\boldsymbol{u}_t = \boldsymbol{\Psi}(\boldsymbol{u}_t^* - \boldsymbol{u}_t)\mathrm{d}t + \boldsymbol{\sigma}\mathrm{d}W_t$*, where* $\boldsymbol{\Psi} = 2\alpha[\boldsymbol{G}^T\boldsymbol{Q}\boldsymbol{G} + \boldsymbol{R}]$*,* $\boldsymbol{\sigma} = \alpha\boldsymbol{B}$ *and* $\boldsymbol{B}^T\boldsymbol{B} = \boldsymbol{\Sigma}_{\boldsymbol{\varepsilon}}$*.*

The proof is provided in Appendix D.

**Theorem 3**: *The stationary distribution of the control recipe searched in Algorithm 2 can be approximated by a multi-normal distribution, which is expressed as*

$$\boldsymbol{u}_t \sim MN\left(\boldsymbol{u}_t^*, \frac{1}{2}\boldsymbol{\sigma}^T\boldsymbol{\Psi}^{-1}\boldsymbol{\sigma}\right), \tag{11}$$

*where* $\boldsymbol{\Psi} = 2\alpha[\boldsymbol{G}^T\boldsymbol{Q}\boldsymbol{G} + \boldsymbol{R}]$ and $\boldsymbol{\sigma} = \alpha\boldsymbol{B}$.

The proof is provided in Appendix E.

In summary, Theorem 1 guarantees the control searched in Algorithm 2 can converge to the underlying optimal one in general. Specifically, if the unknown function $H(\cdot)$ can be approximated by its second-order Taylor expansion, Theorems 2 and 3 propose the explicit formulations of the search process and stationary distribution of control recipes, respectively. Furthermore, from the distribution of control recipes in Equation (11), we find that smaller step sizes can reduce the variations of $\boldsymbol{u}_t$.

## *B. Outer loop: Bayesian inference of disturbances*

In Section 2.1, the prior probability of disturbance $\boldsymbol{d}_t$ is defined as $p(\boldsymbol{d}_t|\boldsymbol{D}_{t-1})$ depending on its trajectory $\boldsymbol{D}_{t-1} = [\boldsymbol{d}_1, \boldsymbol{d}_2, \ldots, \boldsymbol{d}_{t-1}]$. After making control decisions and observing the system output $\boldsymbol{y}_t$, we can update the posterior probability of disturbance $\boldsymbol{d}_t$ using Bayesian inference as follows:

$$p(\boldsymbol{d}_t|\boldsymbol{y}_t) = \frac{p(\boldsymbol{d}_t|\boldsymbol{D}_{t-1})p(\boldsymbol{y}_t|\boldsymbol{d}_t)}{p(\boldsymbol{y}_t)} \propto p(\boldsymbol{d}_t|\boldsymbol{D}_{t-1})p(\boldsymbol{y}_t|\boldsymbol{d}_t), \tag{12}$$

where the conditional probability $p(\boldsymbol{y}_t|\boldsymbol{d}_t)$ can be obtained by Monte Carlo methods based on the system outputs after the convergence of control recipes in Algorithm 2. In literature, the disturbance $\boldsymbol{d}_t$ is generally supposed to be normally distributed given its historical trajectory. Specifically, if $p(\boldsymbol{y}_t|\boldsymbol{d}_t)$ can also be approximated by a normal distribution, we have Proposition 2 for the posterior distribution of the disturbance using Bayesian inference theory as follows.

**Proposition 2**: *If the prior distribution of the disturbance follows multi-normal distribution as* $\boldsymbol{d}_t|\boldsymbol{D}_{t-1}\sim MN(\boldsymbol{\mu}_t, \boldsymbol{\Sigma}_t)$*, the explicit expression of the posterior distribution disturbances after observing the system output* $\boldsymbol{y}_t$ *is given by:*

$$p(\boldsymbol{d}_t|\boldsymbol{y}_t) \propto \exp\left\{-\frac{1}{2}\left(\left(\boldsymbol{y}_t - \frac{1}{N}\sum_{i=1}^{N}\widehat{\boldsymbol{y}}_t(\widehat{\boldsymbol{u}}_t(i))\right)^T \frac{1}{N}\boldsymbol{\Sigma}_y^{-1}\left(\boldsymbol{y}_t - \frac{1}{N}\sum_{i=1}^{N}\widehat{\boldsymbol{y}}_t(\widehat{\boldsymbol{u}}_t(i))\right) + (\boldsymbol{d}_t - \boldsymbol{\mu}_t)^T\boldsymbol{\Sigma}_t^{-1}(\boldsymbol{d}_t - \boldsymbol{\mu}_t)\right)\right\},$$

*where* $\boldsymbol{\Sigma}_y$ *is the sample variance matrix of system output after the convergence of control recipes.*

Notably, other distributions of disturbances can also be updated by Bayesian inference methods using Monte Carlo methods. By analyzing the posterior probability of disturbances, we can obtain a more reliable prior distribution to reduce variations of disturbances in the next run. Algorithm 3 presents the Bayesian update procedure of disturbance as follows.

---

**Algorithm 3. Update distributions of disturbances**

---

**Initialize** $t$, $\boldsymbol{u}_1^{[0]}$, the prior distribution of disturbance $p(\cdot)$, initial disturbance $\boldsymbol{d}_0$.

**For** $t = 1{:}T$

$\boldsymbol{\mu}_t = \int_{-\infty}^{+\infty} \boldsymbol{d}_t \cdot p(\boldsymbol{d}_t|\boldsymbol{D}_{t-1})\mathrm{d}\boldsymbol{d}_t$

$\overline{\boldsymbol{u}}_t \leftarrow \text{Control_Search}(\boldsymbol{\mu}_t)$ **/*Algorithm 2*/**

Take control $\overline{\boldsymbol{u}}_t$, and record the system output $\boldsymbol{y}_t$.

Update the disturbance according to:

$$p(\boldsymbol{d}_t|\boldsymbol{y}_t) = \frac{p(\boldsymbol{d}_t|\boldsymbol{D}_{t-1})p(\boldsymbol{y}_t|\boldsymbol{d}_t)}{p(\boldsymbol{y}_t)} \propto p(\boldsymbol{d}_t|\boldsymbol{D}_{t-1})p(\boldsymbol{y}_t|\boldsymbol{d}_t)$$

Update $p(\boldsymbol{d}_{t+1}|\boldsymbol{D}_t)$.

**End for**

---

## 3.2 *Online control in Phase II*

In real applications of semiconductor manufacturing processes, after control optimization by VM systems in Phase I, real-time control recipes need to be directly determined in practical manufacturing processes. Therefore, in this section, we propose a real-time control algorithm used for online control in Phase II.

Suppose that manufacturing environments and process models are kept stable in Phases I and II, and it is reasonable that the control recipes searched in Phase I can be applied in Phase II. We denote the offline experimental dataset collected in Phase I as $\{D_off\}$. Each sample in $\{D_off\}$ consists of the control recipes, system output, and the distribution of disturbances, i.e., $[\boldsymbol{u}_t, \boldsymbol{y}_t, \boldsymbol{d}_t] \in \{D_off\}$.

Due to the asymptotic optimality of searched control recipes in the offline dataset $\{D_off\}$, it can be used as a "memory buffer" for online control. Since the key hidden variables in manufacturing processes are disturbances, online control decisions can be implemented by matching the closest offline disturbance $\boldsymbol{d}_{t^*}$ in $\{D_off\}$ with the online inferred disturbance and choosing the corresponding control recipe as the online recipe. Specifically, $\boldsymbol{d}_{t^*}$ is obtained by:

$$\boldsymbol{d}_{t^*} \coloneqq arg \min_{\boldsymbol{d} \in \{D_off\}} \mathbb{D}_{KL}(p(\boldsymbol{d}) || q(\boldsymbol{d}_t^{on} | \boldsymbol{D}_{t-1}^{on})), \tag{13}$$

where $\boldsymbol{d}_t^{on}$ is online disturbance and $\mathbb{D}_{KL}(\cdot \,||\, \cdot)$ is Kullback-Leibler divergence. To distinguish the online disturbance, we use $q(\cdot)$ to denote its prior distribution. Then, the control recipe $\boldsymbol{u}_{t^*}$ corresponding to $\boldsymbol{d}_{t^*}$ is chosen as the online control strategy. Notably, as the size of dataset $\{D_off\}$ increases, the divergence between the online and offline disturbance becomes smaller, and the control performs better. Algorithm 4 presents the online control scheme in detail.

**Algorithm 4. Online control in Phase II**

**Input:** Historical offline data $\{D_off\}$, initial system output $y_0$, prior distribution of online disturbance $q(\cdot)$

**For** $t = 1{:}T$

$$\boldsymbol{d}_{t^*} \coloneqq arg \min_{\boldsymbol{d} \in \{D_off\}\}} \mathbb{D}_{KL}[p(\boldsymbol{d}) || q(\boldsymbol{d}_t^{on} | \boldsymbol{D}_{t-1}^{on})]$$

Take the control recipe $\boldsymbol{u}_{t^*}$ corresponding to $\boldsymbol{d}_{t^*}$, and collect the output $\boldsymbol{y}_t$.

Update the disturbance according to $q(\boldsymbol{d}_t^{on}|\boldsymbol{y}_t) = \frac{q(\boldsymbol{d}_t^{on}|\boldsymbol{D}_{t-1}^{on})p(\boldsymbol{y}_t|\boldsymbol{d}_t^{on})}{p(\boldsymbol{y}_t)} \propto$ $q(\boldsymbol{d}_t^{on}|\boldsymbol{D}_{t-1}^{on})p(\boldsymbol{y}_t|\boldsymbol{d}_t^{on})$.

Cauculate $q(\boldsymbol{d}_{t+1}^{on}|\boldsymbol{D}_t^{on})$.

**End for**

# 4. NUMERICAL STUDY AND COMPARISON

To show the performance of the proposed MFRL-BI control scheme, we propose numerical studies based on a nonlinear chemical mechanical planarization (CMP) process in semiconductor manufacturing. In Section 4.1, the proposed MFRL-BI controller is compared with the basic MFRL controller to verify the improvement by using Bayesian inference. In Section 4.2, we focus on a comparison between the proposed MFRL-BI controller and the DOE-based automatic process controller (APC), which is also designed for an unknown process model.

## 4.1 *Comparison with basic MFRL controller*

Due to the privacy of real CMP data, Khuri (1996) proposed an experiment tool and designed a nonlinear process model to describe the CMP process, which is widely used in CMP data simulation (Del Castillo and Yeh, 1998). In this section, we also follow their simulation for data generation. The control recipe $\boldsymbol{u}_t$ consists of three dimensions (i.e., $\boldsymbol{u}_t = \left[u_t^{(1)}, u_t^{(2)}, u_t^{(3)}\right]^{\boldsymbol{T}}$), which represent the backpressure downforce, platen speed, and slurry concentration, respectively. The two dimensions of the system outputs ($\boldsymbol{y}_t = \left[y_t^{(1)}, y_t^{(2)}\right]^{\boldsymbol{T}}$) to reflect the manufacturing quality are removal rate and within-wafer standard deviation with target levels as $\boldsymbol{y}^* = [2200{,}400]^T$. Without loss of generality, the initial system output is set as the target levels.

Specifically, following the nonlinear model proposed by Del Castillo and Yeh (1998), we use the following formulation to simulate data in CMP process at each run $t$.

$$\boldsymbol{y}_t = \boldsymbol{C}\boldsymbol{X}_t + \boldsymbol{d}_t, \tag{14}$$

where $\boldsymbol{C}$ is the parameter matrix defined as

$$\boldsymbol{C} = \begin{bmatrix} 2756.5 & 547.6 & 616.3 & -126.7 & -1109.5 & -286.1 & 989.1 & -52.9 & -156.9 & -550.3 & -10 \\ 746.3 & 62.3 & 128.6 & -152.1 & -289.7 & -32.1 & 237.7 & -28.9 & -122.1 & -140.6 & 1.5 \end{bmatrix},$$

$\boldsymbol{X}_t$ consists of constant, linear, and quadratic terms of control recipes at run $t$

$$\boldsymbol{X}_t = \left[1,\ \ u_t^{(1)},\ \ u_t^{(2)},\ \ u_t^{(3)},\ \ \left[u_t^{(1)}\right]^2,\ \ \left[u_t^{(2)}\right]^2,\ \ \left[u_t^{(3)}\right]^2,\ \ u_t^{(1)}u_t^{(2)},\ \ u_t^{(1)}u_t^{(3)},\ \ u_t^{(2)}u_t^{(3)},\ \ t\right]^{\boldsymbol{T}}.$$

$\boldsymbol{d}_t = \left[d_t^{(1)}, d_t^{(2)}\right]^{\boldsymbol{T}}$ are two dimensions of disturbances that follow two independent IMA(1,1) processes, and the total number of runs $T$ is 50. Based on this setting, we analyze the performance of the proposed MFRL-BI controller and compare it with the basic MFRL controller.

We first consider a special case where there is no extra cost associated with control actions, i.e., $\boldsymbol{R} = \boldsymbol{0}$, the control cost is $C_t(\boldsymbol{u}_t) = (\boldsymbol{y}_t - \boldsymbol{y}^*)^T \boldsymbol{Q} (\boldsymbol{y}_t - \boldsymbol{y}^*)$. Under this setting, the basic MFRL and MFRL-BI controllers are applied for online control, and the corresponding system outputs are used to evaluate the performances of these two controllers. To make a fair comparison, we search control recipes for 2000 iterations at each run in both Algorithms 1 and 2 in the basic MFRL and MFRL-BI controllers, respectively. After collecting data from 1000 production cycles in $\{D_off\}$, we make the online control by matching the disturbances in $\{D_off\}$ with the online one using Algorithm 4. Figure 4 illustrates the boxplot of system outputs in Phase II with 100 replications. The two panels in Figures 4(a) and 4(b) correspond to the two dimensions of $\boldsymbol{y}_t$. As shown, system outputs based on the basic MFRL controller have relatively large variations and significant deviations when dealing with system drifts, while the proposed MFRL-BI controller can keep the system outputs well close to their desired targets, even though the process model is unknown.

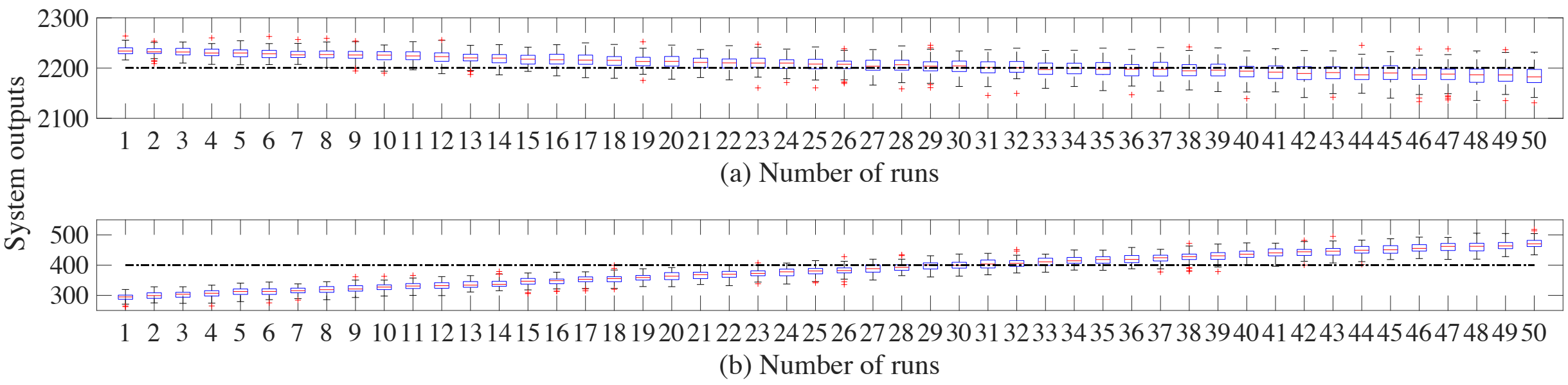

Figure 4(a). Online control results based on the basic MFRL controller

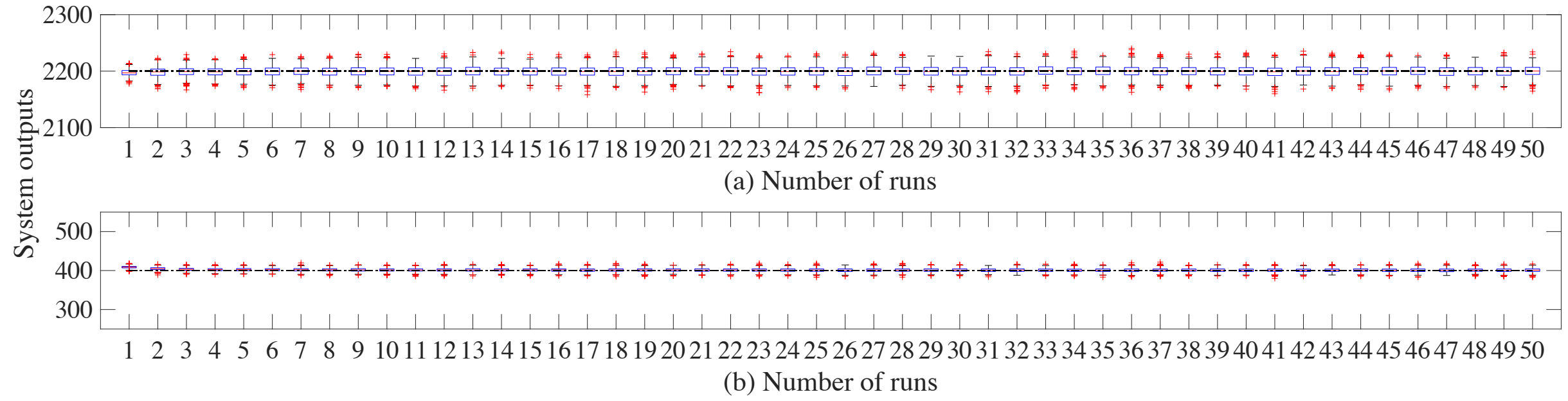

Figure 4(b). Online control results based on the MFRL-BI controller

Generally, executing control has extra control cost during the manufacturing process, the total cost is: $C_t(\boldsymbol{u}_t) = (\boldsymbol{y}_t - \boldsymbol{y}^*)^T \boldsymbol{Q} (\boldsymbol{y}_t - \boldsymbol{y}^*) + \boldsymbol{u}_t^T \boldsymbol{R} \boldsymbol{u}_t$, where $\boldsymbol{R} \neq \boldsymbol{0}$. For example, we set $\boldsymbol{Q} = \begin{bmatrix} 1 & 0 \\ 0 & 1 \end{bmatrix}$ and $\boldsymbol{R} = \begin{bmatrix} 10 & & \\ & 10 & \\ & & 5 \end{bmatrix}$. The mean control cost (MCC) at each run (defined as $\sum_{t=1}^{T} C_t(\boldsymbol{y}_t, \boldsymbol{u}_t) / T$) is used as performance criteria. Table 1 summarizes the mean and standard deviation of MCC in basic MFRL, MFRL-BI controllers, and without control under 100 replications.

Table 1. Comparisons of basic MFRL and MFRL-BI controllers

| Different cases | MCC | Without control | Basic MFRL controller in Algo.1 | MFRL-BI controller in Algo.2-4 |
|---|---|---|---|---|
| $\boldsymbol{R} = \boldsymbol{0}$ | Mean | $2.5989 \times 10^5$ | $3.7054 \times 10^3$ | 116.4702 |
| | *Std.* | $6.9650 \times 10^3$ | 382.4001 | 21.3797 |
| $\boldsymbol{R} \neq \boldsymbol{0}$ | Mean | $2.5989 \times 10^5$ | $5.1766 \times 10^3$ | 135.8367 |
| | *Std.* | $6.9650 \times 10^3$ | 386.9175 | 22.2550 |

As shown in Table 1, in comparison to without control, the basic MFRL controller presented in Algorithm 1 substantially reduces the control cost. Nonetheless, the performance of the basic MFRL does not fulfill the accuracy specifications for semiconductor manufacturing. Upon updating the distribution of disturbances by Algorithms 2 to 4, it is observed that the mean of control cost reduces by 97% in comparison to the basic MFRL controller. Table 1 demonstrates the efficient performance of the MFRL-BI controller in further reducing the control cost during the manufacturing process.

## 4.2 *Comparison with the DOE-based APC*

When process models are unknown, extensive DOE-based methods are proposed in literature for a predictive process model design (Tseng et al., 2019; Shi, 2022). One of the most important methods is the DOE-based automatic process controller (APC) proposed by Zhong et al. (2009), which primarily emphasizes designing experiments to identify the effects of control and disturbances. As the MFRL-BI control scheme also focuses on control optimization based on experimental data when the process model is unknown, we provide a performance comparison with the DOE-based APC. Considering the fairness of performance comparison, we follow the objective of DOE-based APC to minimize the difference between system outputs and their target levels.

### 4.2.1 *Settings of DOE-based APC*

In the methodology of Zhong et al. (2009), the DOE-based APC aims to identify factors that significantly impact system outputs from control recipes, noises in manufacturing environments, and their interactions using a linear DOE regression model. Then, control recipes are optimized considering the randomness of regression parameters. As the nonlinear CMP process is a dynamic manufacturing process with unstable auto-correlated disturbances, current DOE-based APC cannot be directly applied. We incorporate two extra factors in this part: (i) the auto-regression term to describe autocorrelations in disturbances, and (ii) the noises of a linear model to represent the inaccuracy of linear model assumptions. Furthermore, we use the error of system outputs $\boldsymbol{z}_t = \boldsymbol{y}_t - \boldsymbol{y}^*$ as the response variable to simplify the model. In summary, the independent variables to be identified by the DOE regression model are output errors at the last run ($\boldsymbol{z}_{t-1}$), control recipes $\boldsymbol{u}_t$, noises of the linear model at the end of the last run ($\boldsymbol{e}_{t-1}$), the number of runs ($t$), and their interactions.

Before designing experiments, we first run the linear regression model to collect noises ($\boldsymbol{e}_t$), which are used to estimate the model inaccuracy. According to Zhou et al. (2003), a dynamic linear model to describe the manufacturing process is given:

$$\boldsymbol{z}_t = \boldsymbol{\beta}_0 + \boldsymbol{\beta}_1 \boldsymbol{u}_t + \boldsymbol{\beta}_2 \boldsymbol{z}_{t-1} + \boldsymbol{\beta}_3 t + \boldsymbol{e}_t. \tag{15}$$

The noises of the dynamic linear model are calculated by $\boldsymbol{e}_t = \hat{\boldsymbol{z}}_t - \boldsymbol{z}_t$. Then, the effects of the current state ($\boldsymbol{z}_{t-1}$), control recipes ($\boldsymbol{u}_t$), current model noises ($\boldsymbol{e}_{t-1}$) and their interactions are considered in the DOE model as follows:

$$\boldsymbol{z}_t = \boldsymbol{\theta}_0 + \boldsymbol{\theta} \boldsymbol{u}_t + \boldsymbol{\gamma} t + \boldsymbol{\vartheta} \boldsymbol{e}_{t-1} + \boldsymbol{\omega} \boldsymbol{z}_{t-1} + \boldsymbol{\rho} \boldsymbol{u}_t \boldsymbol{e}_{t-1} + \boldsymbol{\varphi} t \boldsymbol{e}_{t-1} + \boldsymbol{r}, \tag{16}$$

where $\boldsymbol{\theta}_0$, $\boldsymbol{\theta}$, $\boldsymbol{\gamma}$, $\boldsymbol{\vartheta}$, $\boldsymbol{\omega}$, $\boldsymbol{\rho}$, and $\boldsymbol{\varphi}$ are the parameter vectors, and $\boldsymbol{r}$ is the residual vector of the DOE model. After selecting the significant variables and their interaction terms by the DOE, we optimize the control recipes as follows:

$$\boldsymbol{u}_t^* = arg \min_{\boldsymbol{u}_t} C_t\left(\boldsymbol{u}_t \middle| \widehat{\boldsymbol{\theta}}_0, \widehat{\boldsymbol{\theta}}, \widehat{\boldsymbol{\gamma}}, \widehat{\boldsymbol{\vartheta}}, \widehat{\boldsymbol{\omega}}, \widehat{\boldsymbol{\rho}}, \widehat{\boldsymbol{\varphi}}, \boldsymbol{e}_{t-1}\right) = arg \min_{\boldsymbol{u}_t} E_{\widehat{\boldsymbol{\theta}}_0, \widehat{\boldsymbol{\theta}}, \widehat{\boldsymbol{\gamma}}, \widehat{\boldsymbol{\vartheta}}, \widehat{\boldsymbol{\omega}}, \widehat{\boldsymbol{\rho}}, \widehat{\boldsymbol{\varphi}}}\left(\boldsymbol{z}_t^T \boldsymbol{z}_t \middle| \widehat{\boldsymbol{\theta}}_0, \widehat{\boldsymbol{\theta}}, \widehat{\boldsymbol{\gamma}}, \widehat{\boldsymbol{\vartheta}}, \widehat{\boldsymbol{\omega}}, \widehat{\boldsymbol{\rho}}, \widehat{\boldsymbol{\varphi}}, \boldsymbol{e}_{t-1}\right) \tag{17}$$

Generally, the DOE-based APC aims to approximate the manufacturing process by a linear regression model, which is unbiased when the ground truth of the process model is linear. However, in this section, we focus mainly on a complex nonlinear CMP process, wherein an exhaustive comparison of DOE-based APC and the proposed MFRL-BI controller are presented.

### 4.2.2 *Numerical comparison*

Numerical comparison results are discussed in this part. For DOE-based APC, we first collect the model noises in Equation (15) using the offline data, which are generated by nonlinear CMP simulations in Equation (14) 1000 times from run 1 to $T$. Then based on Equation (16), the effects of control variables ($\boldsymbol{u}_t$), the number of runs ($t$), and the model noises ($\boldsymbol{e}_{t-1}$) on the response variable ($\boldsymbol{z}_t$) are summarized in Table 2. Specifically, $z_t^{(1)}$ ($e_t^{(1)}$) and $z_t^{(2)}$ ($e_t^{(2)}$) denote the two dimensions of the response variables (noises) at run $t$. The experiments in each cell are replicated 300 times to calculate the mean response values of $\boldsymbol{z}_t$.

Table 2. Design and responses for the Nonlinear CMP modeling experiment

| **Control variables** | | | | | **Response variable $\boldsymbol{z}_t^{(1)}$ for noises $\boldsymbol{e}_{t-1}^{(1)}$** | | **Response variable $\boldsymbol{z}_t^{(2)}$ for noises $\boldsymbol{e}_{t-1}^{(2)}$** | |
|---|---|---|---|---|---|---|---|---|
| Cell | $u_t^{(1)}$ | $u_t^{(2)}$ | $u_t^{(3)}$ | $t$ | − | + | − | + |
| 1 | − | − | − | − | 434.74 | 419.06 | 369.77 | 362.09 |
| 2 | − | + | + | − | 1077.58 | 1060.30 | 411.90 | 403.40 |
| 3 | + | + | − | − | 149.25 | 133.07 | 209.13 | 202.49 |
| 4 | + | − | + | − | 576.23 | 561.05 | 106.09 | 98.13 |
| 5 | − | + | − | + | 512.51 | 501.30 | 501.25 | 495.98 |
| 6 | − | − | + | + | 1041.75 | 1031.02 | 492.37 | 486.61 |
| 7 | + | − | − | + | -380.48 | -390.94 | 179.42 | 173.98 |
| 8 | + | + | + | + | 52.38 | 44.81 | 68.80 | 63.02 |

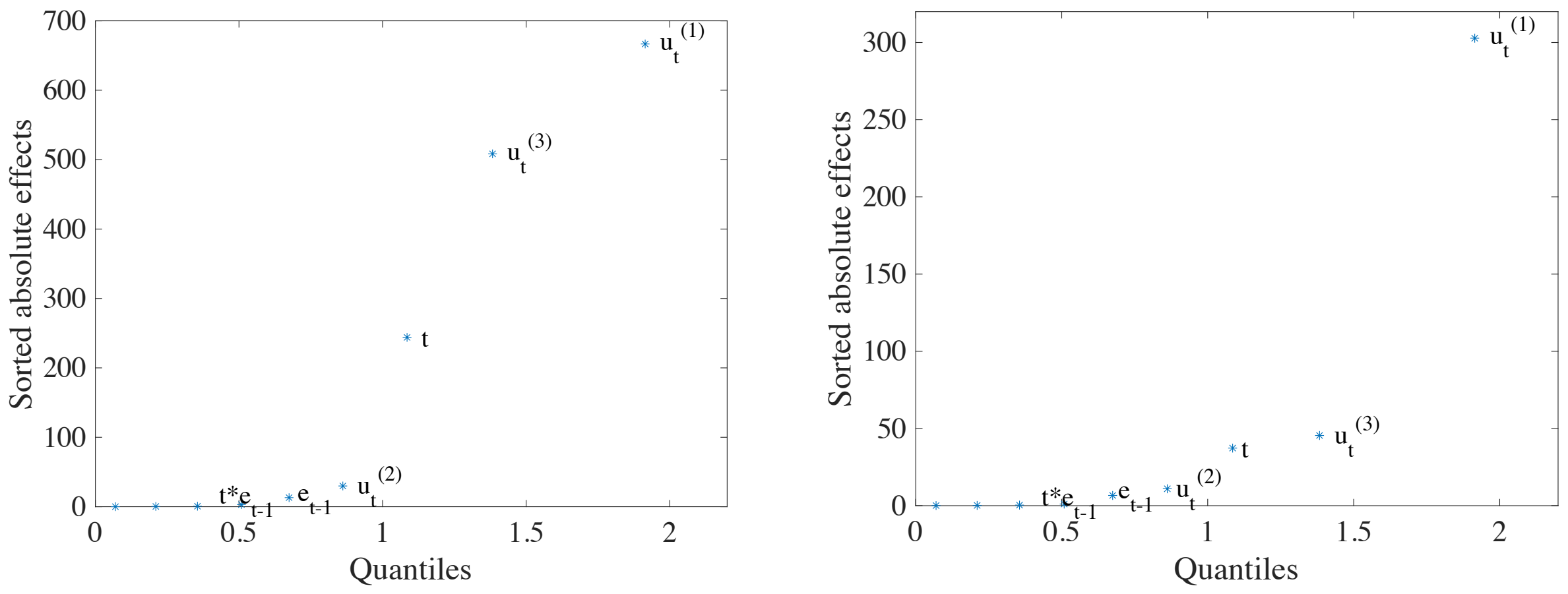

(a) $z_t^{(1)}$ (b) $z_t^{(2)}$

Figure 5. Half-normal probability plot of main effects and interactions

By illustrating the main effects and their interactions using the half-normal plot in Figure 5, we identify the significant terms and obtain the DOE-based approximate model as follows:

$$\begin{cases} z_t^{(1)} = \theta_{10} + \theta_{11}u_t^{(1)} + \theta_{12}u_t^{(2)} + \theta_{13}u_t^{(3)} + \gamma_1 t + \vartheta_1 e_{t-1}^{(1)} + \omega_1 z_{t-1}^{(1)} + \varphi_1 t e_{t-1}^{(1)} + r_1 \\ z_t^{(2)} = \theta_{20} + \theta_{21}u_t^{(1)} + \theta_{22}u_t^{(2)} + \theta_{23}u_t^{(3)} + \gamma_2 t + \vartheta_2 e_{t-1}^{(2)} + \omega_2 z_{t-1}^{(2)} + \varphi_2 t e_{t-1}^{(2)} + r_2. \end{cases} \tag{18}$$

We define $\boldsymbol{\theta_0} = \begin{bmatrix}\theta_{10}\\ \theta_{20}\end{bmatrix}$, $\boldsymbol{\theta} = \begin{bmatrix}\boldsymbol{\theta_1}\\ \boldsymbol{\theta_2}\end{bmatrix} = \begin{bmatrix}\theta_{11} & \theta_{12} & \theta_{13}\\ \theta_{21} & \theta_{22} & \theta_{23}\end{bmatrix}$, $\boldsymbol{\gamma} = \begin{bmatrix}\gamma_1\\ \gamma_2\end{bmatrix}$, $\boldsymbol{\vartheta} = \begin{bmatrix}\vartheta_1\\ \vartheta_2\end{bmatrix}$,$\boldsymbol{\omega} = \begin{bmatrix}\omega_1\\ \omega_2\end{bmatrix}$, $\boldsymbol{\varphi} = \begin{bmatrix}\varphi_1\\ \varphi_2\end{bmatrix}$ as parameters that need to be estimated. Due to the randomness of $\boldsymbol{r} = [r_1, r_2]^T$ in Equation (18), the parameter estimators $\widehat{\boldsymbol{\theta}}_{\mathbf{0}}$, $\widehat{\boldsymbol{\theta}}$, $\widehat{\boldsymbol{\gamma}}$, $\widehat{\boldsymbol{\vartheta}}$, $\widehat{\boldsymbol{\omega}}$, and $\widehat{\boldsymbol{\varphi}}$ are also random variables. Moreover, $\boldsymbol{e}_t$ is used to describe the model noise in Equation (15), which is also a random vector. Figures 6(a) and 6(b) display the distribution of model noises and parameter estimators, respectively. Subsequently, a robust control recipe considering the randomness of variables in Figure 6 is designed with a closed-form solution according to Zhong et al. (2009) (see Appendix F for more detailed derivations).

$$\boldsymbol{u_t^*} = arg\min_{\boldsymbol{u_t}} E_{\widehat{\boldsymbol{\theta}}_0,\widehat{\boldsymbol{\theta}},\widehat{\boldsymbol{\gamma}},\widehat{\boldsymbol{\vartheta}},\widehat{\boldsymbol{\omega}},\widehat{\boldsymbol{\rho}},\widehat{\boldsymbol{\varphi}}}\left(\boldsymbol{z_t^T z_t} \middle| \widehat{\boldsymbol{\theta}}_{\mathbf{0}}, \widehat{\boldsymbol{\theta}}, \widehat{\boldsymbol{\gamma}}, \widehat{\boldsymbol{\vartheta}}, \widehat{\boldsymbol{\omega}}, \widehat{\boldsymbol{\varphi}}, \boldsymbol{e}_{t-1}\right) = -\left[\Sigma_{\boldsymbol{\theta}}^{\mathbf{1}} + \widehat{\boldsymbol{\theta}}_{\mathbf{1}}\widehat{\boldsymbol{\theta}}_{\mathbf{1}}^{\boldsymbol{T}} + \Sigma_{\boldsymbol{\theta}}^{\mathbf{2}} + \widehat{\boldsymbol{\theta}}_{\mathbf{2}}\widehat{\boldsymbol{\theta}}_{\mathbf{2}}^{\boldsymbol{T}}\right]^{-\mathbf{1}}$$
$$\cdot \left[\left(\hat{\theta}_{10} + \hat{\gamma}_1 t + \hat{\vartheta}_1 e_{t-1}^{(1)} + \hat{\varphi}_1 t e_{t-1}^{(1)} + \hat{\omega}_1 z_{t-1}^{(1)}\right) \cdot \widehat{\boldsymbol{\theta}}_{\mathbf{1}} + \left(\hat{\theta}_{20} + \hat{\gamma}_2 t + \hat{\vartheta}_2 e_{t-1}^{(2)} + \hat{\varphi}_2 t e_{t-1}^{(2)} + \hat{\omega}_2 z_{t-1}^{(2)}\right) \cdot \widehat{\boldsymbol{\theta}}_{\mathbf{2}}\right], \tag{19}$$

where $\widehat{\boldsymbol{\theta}}_{\mathbf{1}} = \left[\hat{\theta}_{11}, \hat{\theta}_{12}, \hat{\theta}_{13}\right]^T$ and $\widehat{\boldsymbol{\theta}}_{\mathbf{2}} = \left[\hat{\theta}_{21}, \hat{\theta}_{22}, \hat{\theta}_{23}\right]^T$. $\Sigma_{\boldsymbol{\theta}}^{\mathbf{1}}$ and $\Sigma_{\boldsymbol{\theta}}^{\mathbf{2}}$ are covariance matrices of $\widehat{\boldsymbol{\theta}}_{\mathbf{1}}$ and $\widehat{\boldsymbol{\theta}}_{\mathbf{2}}$ respectively.

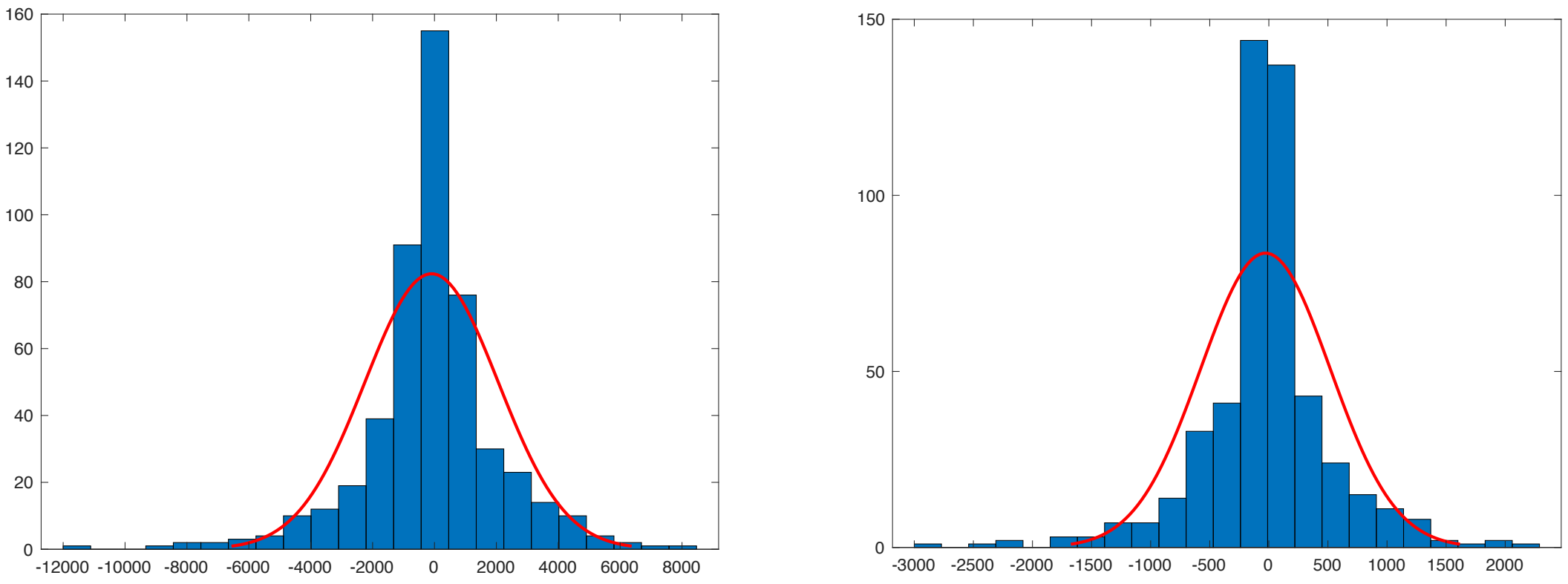

Figure 6(a). Histogram of noises in the dynamic linear model ($\boldsymbol{e}_t$)

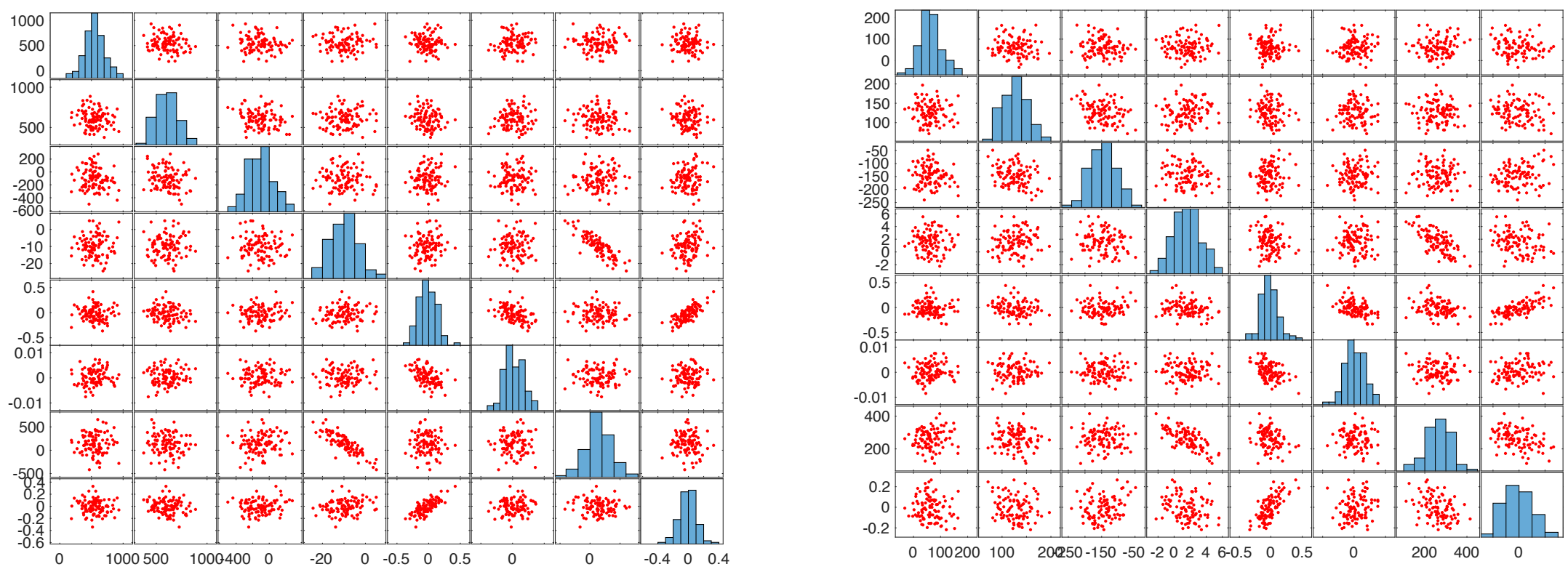

Figure 6(b). Distribution of $\hat{\theta}_{10}, \hat{\boldsymbol{\theta}}_{\mathbf{1}}, \hat{\gamma}_1, \hat{\vartheta}_1, \hat{\varphi}_1, \hat{\omega}_1$ and $\hat{\theta}_{20}, \hat{\boldsymbol{\theta}}_{\mathbf{2}}, \hat{\gamma}_2, \hat{\vartheta}_2, \hat{\varphi}_2, \hat{\omega}_2$

To make a fair comparison, we employ the same amount of historical data in the MFRL-BI controller and DOE-based APC. However, it is difficult for a linear DOE-based regression model to approximate a nonlinear CMP process. As shown in Figure 7, according to the closed-form solution in Equation (19), DOE-based APC even cannot keep the system outputs close to the desired target. When compared with the MFRL-BI controller in Figure 4(b), the linear DOE-based APC is invalid when the underlying process model is nonlinear. Table 3 presents the mean and standard deviation of MCC based on the MFRL-BI controller and DOE-based APC under 100 replications. The results demonstrate that the MFRL-BI controller surpasses the DOE-based APC in nonlinear CMP processes, implying that the proposed MFRL-BI controller can overcome the limitations of linear DOE-based APC when dealing with more complex nonlinear processes.

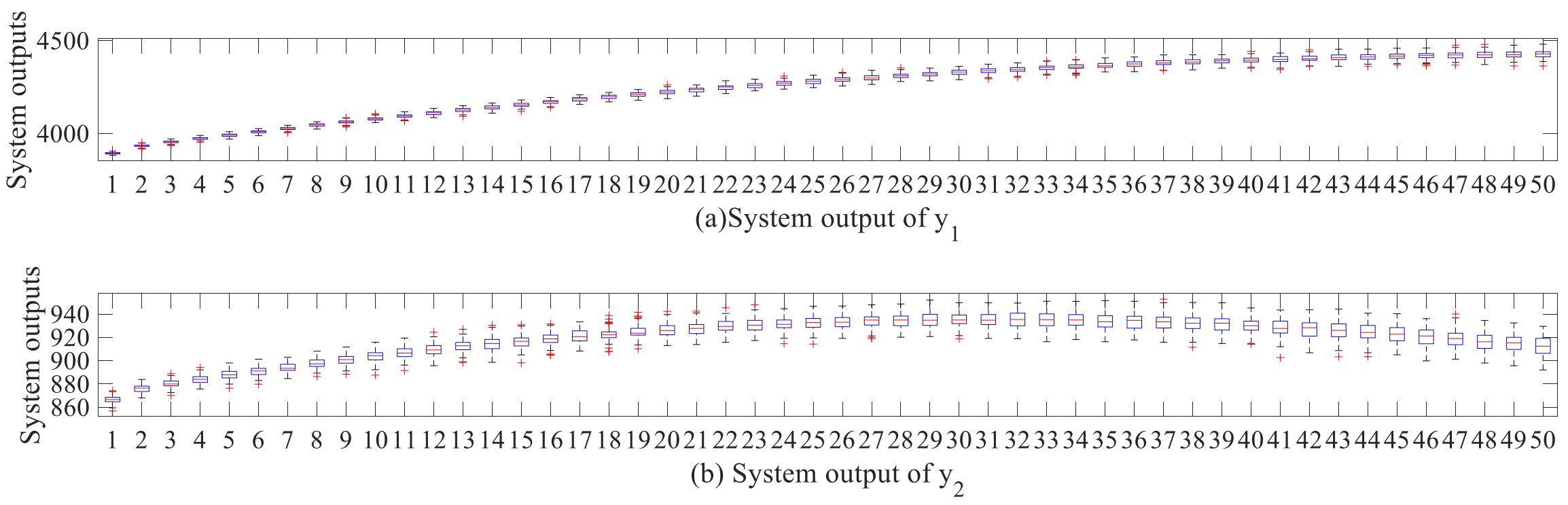

Figure 7. Control results based on linear DOE-based APC

Table 3. MCC of MFRL-BI controller and DOE-based APC

| Controllers | Mean of MCC | Std. of MCC |
|---|---|---|
| MFRL-BI controller | 116.4702 | 21.3797 |
| DOE-based APC | $4.5408 \times 10^6$ | $4.5314 \times 10^4$ |

# 5. CONCLUSIONS

This work designs a new process control scheme by model-free reinforcement learning to reduce the system variations in semiconductor manufacturing when process model is unknown and complex. Due to unstable and unobservable disturbances, basic MFRL controllers usually suffer from large variations. To overcome this challenge, We update the distribution of disturbances during manufacturing processes using Bayesian inference. The corresponding algorithms in offline optimization and online control phases are presented, and corresponding theoretical properties are also guaranteed. Through performance comparisons between the proposed MFRL-BI, basic MFRL, and DOE-based APC, we observe that the proposed MFRL-BI controller exhibits superior performance, particularly when underlying process models are nonlinear and complex.

Along with our research direction, several extensions can be further investigated. First, how to develop a RL-based process control model when the effects of control recipes and disturbances are correlated. Second, the constraints of control recipes can also be considered in process control optimization in future studies.

# APPENDIX

## *Appendix A. Simplification of the optimization problem in Equation (7)*

The optimization problem in Equation (7) can be simplified as follows:

$$
\begin{aligned}
\mathbf{E}_{\boldsymbol{\delta}_t,\boldsymbol{v}_t}[C_t(\boldsymbol{y}_t,\boldsymbol{u}_t)] &= \mathbf{E}_{\boldsymbol{\delta}_t,\boldsymbol{v}_t}[(\boldsymbol{y}_t-\boldsymbol{y}^*)^T\boldsymbol{Q}(\boldsymbol{y}_t-\boldsymbol{y}^*)+\boldsymbol{u}_t^T\boldsymbol{R}\boldsymbol{u}_t] \\
&= \mathbf{E}_{\boldsymbol{\delta}_t,\boldsymbol{v}_t}[(g(\boldsymbol{u}_t)+\boldsymbol{d}_t-\boldsymbol{y}^*)^T\boldsymbol{Q}(g(\boldsymbol{u}_t)+\boldsymbol{d}_t-\boldsymbol{y}^*)]+\boldsymbol{u}_t^T\boldsymbol{R}\boldsymbol{u}_t \\
&= \mathbf{E}_{\boldsymbol{v}_t}\left[\mathbf{E}_{\boldsymbol{\delta}_t}[(g(\boldsymbol{u}_t)+\boldsymbol{\mu}_t+\boldsymbol{\delta}_t-\boldsymbol{y}^*)^T\boldsymbol{Q}(g(\boldsymbol{u}_t)+\boldsymbol{\mu}_t+\boldsymbol{\delta}_t-\boldsymbol{y}^*)]\right]+\boldsymbol{u}_t^T\boldsymbol{R}\boldsymbol{u}_t \\
&= \mathbf{E}_{\boldsymbol{v}_t}[tr(\boldsymbol{Q}\boldsymbol{\Sigma}_t)+(g(\boldsymbol{u}_t)+\boldsymbol{\mu}_t-\boldsymbol{y}^*)^T\boldsymbol{Q}(g(\boldsymbol{u}_t)+\boldsymbol{\mu}_t-\boldsymbol{y}^*)]+\boldsymbol{u}_t^T\boldsymbol{R}\boldsymbol{u}_t \\
&= tr(\boldsymbol{Q}\boldsymbol{\Sigma}_t)+\mathbf{E}_{\boldsymbol{v}_t}[(g(\boldsymbol{u}_t)+\boldsymbol{\mu}_t-\boldsymbol{y}^*)^T\boldsymbol{Q}(g(\boldsymbol{u}_t)+\boldsymbol{\mu}_t-\boldsymbol{y}^*)]+\boldsymbol{u}_t^T\boldsymbol{R}\boldsymbol{u}_t.
\end{aligned}
$$

This result is directly presented in Equation (8).

## *Appendix B. Proof of Theorem 1*

As we analyzed in Section 3.2, the iteration of control recipes is defined as $\Delta\boldsymbol{u}_t = -\alpha\nabla_{\boldsymbol{u}_t}M(\boldsymbol{u}_t|\boldsymbol{\mu}_t)$. By combining Assumptions 3.1 and 3.2, we have

$$\mathrm{d}\boldsymbol{u}_t = -\boldsymbol{\alpha}\nabla_{\boldsymbol{u}_t}M(\boldsymbol{u}_t|\boldsymbol{\mu}_t)\mathrm{d}t+\boldsymbol{\alpha}\boldsymbol{B}\mathrm{d}W_t, \tag{B.1}$$

where $\boldsymbol{B}^{\boldsymbol{T}}\boldsymbol{B}=\boldsymbol{\Sigma}_\varepsilon$, and $W_t$ is a standard Wiener process.

Based on Equation (B.1), the control action search process can be approximated by a Fokker-Planck equation, which has a standard expression: $\boldsymbol{du}_t = A(\boldsymbol{u}_t,t)dt+\left(B(\boldsymbol{u}_t,t)\right)^{1/2}\mathrm{d}W_t$. In our work, we have $A(\boldsymbol{u}_t,t)=-\alpha\nabla_{\boldsymbol{u}_t}M(\boldsymbol{u}_t,\boldsymbol{\mu}_t)$ and $B(\boldsymbol{u}_t,t)=(\alpha\boldsymbol{B})^T\alpha\boldsymbol{B}$. We find that $B(\boldsymbol{u}_t,t)$ which is a constant matrix that is independent with $\boldsymbol{u}_t$. According to Gardiner (1985), $\boldsymbol{u}_t$ has a stable distribution $p_s(\boldsymbol{u}_t)$ if

$$\nabla\boldsymbol{u}_t[A(\boldsymbol{u}_t,t)p_s(\boldsymbol{u}_t)]-\frac{1}{2}\nabla^2_{\boldsymbol{u}_t}[B(\boldsymbol{u}_t,t)p_s(\boldsymbol{u}_t)]=0. \tag{B.2}$$

We can find the stable distribution as

$$p_s(\boldsymbol{u}_t)\propto e^{-\frac{2\alpha M(\boldsymbol{u}_t|\boldsymbol{\mu}_t)}{(\alpha\boldsymbol{B})^T\alpha\boldsymbol{B}}}=\exp\left\{-\frac{2\alpha[(g(\boldsymbol{u}_t)+\boldsymbol{\mu}_t-\boldsymbol{y}^*)^T\boldsymbol{Q}(g(\boldsymbol{u}_t)+\boldsymbol{\mu}_t-\boldsymbol{y}^*)+\boldsymbol{u}_t^T\boldsymbol{R}\boldsymbol{u}_t]}{(\alpha\boldsymbol{B})^T\alpha\boldsymbol{B}}\right\}. \tag{B.3}$$

Therefore, we can conclude that the control recipe obtained by Algorithm 2 has a stable distribution with mean vector $\mathbf{E}[\boldsymbol{u}_t]=\boldsymbol{u}_t^*\coloneqq\arg\min_{\boldsymbol{u}_t}(g(\boldsymbol{u}_t)+\boldsymbol{\mu}_t-\boldsymbol{y}^*)^T\boldsymbol{Q}(g(\boldsymbol{u}_t)+\boldsymbol{\mu}_t-\boldsymbol{y}^*)+\boldsymbol{u}_t^T\boldsymbol{R}\boldsymbol{u}_t$.

## *Appendix C. Proof of Proposition 1*

If $H(\boldsymbol{u}_t|\boldsymbol{\mu}_t)$ can be approximated as its second-order Taylor expansion, we have the Taylor expansion of $H(\boldsymbol{u}_t|\boldsymbol{\mu}_t)$ at point $\tilde{\boldsymbol{u}}_t$ as:

$$\begin{aligned} H(\boldsymbol{u}_t|\boldsymbol{\mu}_t) \quad &\approx H(\tilde{\boldsymbol{u}}_t|\boldsymbol{\mu}_t) + \nabla_{\boldsymbol{u}_t} H(\boldsymbol{u}_t|\boldsymbol{\mu}_t)|_{\boldsymbol{u}_t=\tilde{\boldsymbol{u}}_t}(\boldsymbol{u}_t - \tilde{\boldsymbol{u}}_t) \\ &+\frac{1}{2}(\boldsymbol{u}_t - \tilde{\boldsymbol{u}}_t)^T \nabla^2_{\tilde{\boldsymbol{u}}_t} H(\boldsymbol{u}_t, \boldsymbol{\mu}_t)|_{\boldsymbol{u}_t=\tilde{\boldsymbol{u}}_t}(\boldsymbol{u}_t - \tilde{\boldsymbol{u}}_t), \end{aligned} \tag{C.1}$$

where $\nabla^2_{\tilde{\boldsymbol{u}}_t} H(\boldsymbol{u}_t|\boldsymbol{\mu}_t)|_{\boldsymbol{u}_t=\tilde{\boldsymbol{u}}_t}$ is Hessian matrix of function $H(\cdot)$ when $\boldsymbol{u}_t$ equals to $\tilde{\boldsymbol{u}}_t$. Then the gradient of $\nabla_{\boldsymbol{u}_t} H(\boldsymbol{u}_t|\boldsymbol{\mu}_t)$ can be approximated as

$$\nabla_{\boldsymbol{u}_t}\left[H(\tilde{\boldsymbol{u}}_t|\boldsymbol{\mu}_t) + \nabla_{\boldsymbol{u}_t} H(\boldsymbol{u}_t|\boldsymbol{\mu}_t)|_{\boldsymbol{u}_t=\tilde{\boldsymbol{u}}_t}(\boldsymbol{u}_t - \tilde{\boldsymbol{u}}_t) + \frac{1}{2}(\boldsymbol{u}_t - \tilde{\boldsymbol{u}}_t)^T \nabla^2_{\tilde{\boldsymbol{u}}_t} H(\boldsymbol{u}_t|\boldsymbol{\mu}_t)|_{\boldsymbol{u}_t=\tilde{\boldsymbol{u}}_t}(\boldsymbol{u}_t - \tilde{\boldsymbol{u}}_t)\right]. \tag{C.2}$$

Since $\tilde{\boldsymbol{u}}_t \coloneqq arg \min\limits_{\boldsymbol{u}_t} H(\boldsymbol{u}_t|\boldsymbol{\mu}_t)$, we have $\nabla_{\boldsymbol{u}_t} H(\boldsymbol{u}_t|\boldsymbol{\mu}_t)|_{\boldsymbol{u}_t=\tilde{\boldsymbol{u}}_t} = 0$. Moreover, $H(\tilde{\boldsymbol{u}}_t|\boldsymbol{\mu}_t)$ is a constant for $\boldsymbol{u}_t$. We only analyze the last term in Equation (C.2). According to the definition of function $H(\cdot)$, we have $\nabla^2_{\tilde{\boldsymbol{u}}_t} H(\boldsymbol{u}_t|\boldsymbol{\mu}_t)|_{\boldsymbol{u}_t=\tilde{\boldsymbol{u}}_t} = 2\boldsymbol{G^T Q G}$ where $\boldsymbol{G} = \begin{bmatrix} \frac{\partial g_1}{\partial \tilde{u}_1} & \cdots & \frac{\partial g_1}{\partial \tilde{u}_m} \\ \vdots & \ddots & \vdots \\ \frac{\partial g_n}{\partial \tilde{u}_1} & \cdots & \frac{\partial g_n}{\partial \tilde{u}_m} \end{bmatrix}_{n\times m}$ is the gradient of multivariate function $g(\cdot)$, and $g_1$ to $g_n$ correspond to $n$ dimensions of $\boldsymbol{y}_t$. Then, based on Equation (C.2), we have $\nabla_{\boldsymbol{u}_t} H(\boldsymbol{u}_t|\boldsymbol{\mu}_t) = 2\boldsymbol{G^T Q G}(\boldsymbol{u}_t - \tilde{\boldsymbol{u}}_t)$. According to the definition of $\boldsymbol{u}_t^*$, we have:

$$\nabla_{\boldsymbol{u}_t} M(\boldsymbol{u}_t^*|\boldsymbol{\mu}_t) = \nabla_{\boldsymbol{u}_t} H(\boldsymbol{u}_t^*|\boldsymbol{\mu}_t) + 2\boldsymbol{R}\boldsymbol{u}_t^* = 2\boldsymbol{G^T Q G}(\boldsymbol{u}_t^* - \tilde{\boldsymbol{u}}_t) + 2\boldsymbol{R}\boldsymbol{u}_t^* = 0.$$

Thus we have $\boldsymbol{u}_t^* = (\boldsymbol{G^T Q G} + \boldsymbol{R})^{-1}\boldsymbol{G^T Q G}\tilde{\boldsymbol{u}}_t$.

## *Appendix D. Proof of Theorem 2*

If $H(\boldsymbol{u}_t|\boldsymbol{\mu}_t)$ can be approximated as its second-order Taylor expansion, we have

$$H(\boldsymbol{u}_t|\boldsymbol{\mu}_t) \approx H(\tilde{\boldsymbol{u}}_t|\boldsymbol{\mu}_t) + \nabla_{\boldsymbol{u}_t} H^T(\boldsymbol{u}_t|\boldsymbol{\mu}_t)|_{\boldsymbol{u}_t=\tilde{\boldsymbol{u}}_t}(\boldsymbol{u}_t - \tilde{\boldsymbol{u}}_t) + \frac{1}{2}(\boldsymbol{u}_t - \tilde{\boldsymbol{u}}_t)^T \nabla^2_{\tilde{\boldsymbol{u}}_t} H(\boldsymbol{u}_t|\boldsymbol{\mu}_t)|_{\boldsymbol{u}_t=\tilde{\boldsymbol{u}}_t}(\boldsymbol{u}_t - \tilde{\boldsymbol{u}}_t). \tag{D.1}$$

Therefore, we have the control action searching process for $\boldsymbol{u}_t^*$ as

$$\begin{aligned} \mathrm{d}\boldsymbol{u}_t \quad &= -\alpha \nabla_{\boldsymbol{u}_t} M(\boldsymbol{u}_t|\boldsymbol{\mu}_t)\mathrm{d}t + \alpha \boldsymbol{B}\mathrm{d}W_t \\ &-\alpha \nabla_{\boldsymbol{u}_t}\big(H(\boldsymbol{u}_t|\boldsymbol{\mu}_t) + \boldsymbol{u}_t^T \boldsymbol{R}\boldsymbol{u}_t\big)\mathrm{d}t + \alpha \boldsymbol{B}\mathrm{d}W_t \\ &\approx -\alpha[2\boldsymbol{G^T Q G}(\boldsymbol{u}_t - \tilde{\boldsymbol{u}}_t) + 2\boldsymbol{R}\boldsymbol{u}_t]\mathrm{d}t + \alpha \boldsymbol{B}\mathrm{d}W_t \\ &= -2\alpha[\boldsymbol{G^T Q G}\boldsymbol{u}_t - \boldsymbol{G^T Q G}\tilde{\boldsymbol{u}}_t + \boldsymbol{R}\boldsymbol{u}_t]\mathrm{d}t + \alpha \boldsymbol{B}\mathrm{d}W_t \\ &= -2\alpha[\boldsymbol{G^T Q G}\boldsymbol{u}_t - (\boldsymbol{G^T Q G} + \boldsymbol{R})\boldsymbol{u}_t^* + \boldsymbol{R}\boldsymbol{u}_t]\mathrm{d}t + \alpha \boldsymbol{B}\mathrm{d}W_t \\ &= 2\alpha\big(\boldsymbol{G^T Q G} + \boldsymbol{R}\big)(\boldsymbol{u}_t^* - \boldsymbol{u}_t)\mathrm{d}t + \alpha \boldsymbol{B}\mathrm{d}W_t. \end{aligned} \tag{D.2}$$

Let $\boldsymbol{\Psi} = 2\alpha[\boldsymbol{G}^T \boldsymbol{Q}\boldsymbol{G} + \boldsymbol{R}]$ and $\boldsymbol{\sigma} = \alpha\boldsymbol{B}$. Because $\alpha$ is a positive step size, we have $\boldsymbol{\Psi} > 0$. Then we have the search process of control action $\boldsymbol{u}_t^*$ is an Ornstein–Uhlenbeck process.

## Appendix E. Proof of Theorem 3

According to Theorem 2, we have the searching process for $\boldsymbol{u}_t^*$ follows Ornstein–Uhlenbeck process, i.e. $\mathrm{d}\boldsymbol{u}_t = \boldsymbol{\Psi}(\boldsymbol{u}_t^* - \boldsymbol{u}_t)\mathrm{d}t + \boldsymbol{\sigma}\mathrm{d}W_t$. According to Gardiner (1985), the stationary distribution of $\boldsymbol{u}_t^*$ satisfies:

$$\nabla_{\boldsymbol{u}_t}[\boldsymbol{\Psi}(\boldsymbol{u}_t^* - \boldsymbol{u}_t)p_s(\boldsymbol{u}_t)|\boldsymbol{u}_t = \boldsymbol{u}_t^*] = \frac{\mathbf{1}}{\mathbf{2}}\boldsymbol{\sigma}^T\boldsymbol{\sigma}\nabla_{\boldsymbol{u}_t}^2[p_s(\boldsymbol{u}_t)|\boldsymbol{u}_t = \boldsymbol{u}_t^*]. \tag{E.1}$$

We can solve the stationary distribution of $\boldsymbol{u}_t$ as: $p_s(\boldsymbol{u}_t) \propto$ is to say, as $t \to \infty$, the search process for $\boldsymbol{u}_t^*$ has a stationary distribution as:

$$\boldsymbol{u}_t \sim MN\left(\boldsymbol{u}_t^*, \frac{1}{2}\boldsymbol{\sigma}^T\boldsymbol{\Psi}^{-1}\boldsymbol{\sigma}\right). \tag{E.2}$$

## Appendix F. Solution of the DOE-based APC

According to Equation (19), the control decision is $\boldsymbol{u}_t^* = arg \min_{\boldsymbol{u}_t} C_t\left(\boldsymbol{u}_t \middle| \widehat{\boldsymbol{\theta}}_{\mathbf{0}}, \widehat{\boldsymbol{\theta}}, \widehat{\boldsymbol{\gamma}}, \widehat{\boldsymbol{\vartheta}}, \widehat{\boldsymbol{\omega}}, \widehat{\boldsymbol{\rho}}, \widehat{\boldsymbol{\varphi}}, \boldsymbol{e}_{t-1}\right) = arg \min_{\boldsymbol{u}_t} \mathbf{E}_{\widehat{\boldsymbol{\theta}}_{\mathbf{0}}, \widehat{\boldsymbol{\theta}}, \widehat{\boldsymbol{\gamma}}, \widehat{\boldsymbol{\vartheta}}, \widehat{\boldsymbol{\omega}}, \widehat{\boldsymbol{\rho}}, \widehat{\boldsymbol{\varphi}}, \boldsymbol{e}_{t-1}}\left(\boldsymbol{z}_t^T\boldsymbol{z}_t \middle| \widehat{\boldsymbol{\theta}}_{\mathbf{0}}, \widehat{\boldsymbol{\theta}}, \widehat{\boldsymbol{\gamma}}, \widehat{\boldsymbol{\vartheta}}, \widehat{\boldsymbol{\omega}}, \widehat{\boldsymbol{\rho}}, \widehat{\boldsymbol{\varphi}}, \boldsymbol{e}_{t-1}\right)$. Since we have $\exp\{-(\boldsymbol{u}_t - \boldsymbol{u}_t^*)^T[\boldsymbol{\sigma}^T\boldsymbol{\Psi}^{-1}\boldsymbol{\sigma}]^{-1}(\boldsymbol{u}_t - \boldsymbol{u}_t^*)\}$. That a two-dimension system output in the CMP case, Equation (19) can be rewritten as

$$\begin{aligned}\boldsymbol{u}_t^* &= arg \min_{\boldsymbol{u}_t} \mathbf{E}_{\widehat{\boldsymbol{\theta}}_{\mathbf{0}}, \widehat{\boldsymbol{\theta}}, \widehat{\boldsymbol{\gamma}}, \widehat{\boldsymbol{\vartheta}}, \widehat{\boldsymbol{\omega}}, \widehat{\boldsymbol{\rho}}, \widehat{\boldsymbol{\varphi}}, \boldsymbol{e}_{t-1}}\left(\left(y_t^{(1)} - y_1^*\right)^2 + \left(y_t^{(2)} - y_2^*\right)^2 \middle| \widehat{\boldsymbol{\theta}}_{\mathbf{0}}, \widehat{\boldsymbol{\theta}}, \widehat{\boldsymbol{\gamma}}, \widehat{\boldsymbol{\vartheta}}, \widehat{\boldsymbol{\omega}}, \widehat{\boldsymbol{\rho}}, \widehat{\boldsymbol{\varphi}}, \boldsymbol{e}_{t-1}\right)\\ &= arg \min_{\boldsymbol{u}_t} \mathbf{E}_{\widehat{\boldsymbol{\theta}}_{\mathbf{0}}, \widehat{\boldsymbol{\theta}}, \widehat{\boldsymbol{\gamma}}, \widehat{\boldsymbol{\vartheta}}, \widehat{\boldsymbol{\omega}}, \widehat{\boldsymbol{\rho}}, \widehat{\boldsymbol{\varphi}}, \boldsymbol{e}_{t-1}}\left(\left(z_t^{(1)}\right)^2 + \left(z_t^{(2)}\right)^2 \middle| \widehat{\boldsymbol{\theta}}_{\mathbf{0}}, \widehat{\boldsymbol{\theta}}, \widehat{\boldsymbol{\gamma}}, \widehat{\boldsymbol{\vartheta}}, \widehat{\boldsymbol{\omega}}, \widehat{\boldsymbol{\rho}}, \widehat{\boldsymbol{\varphi}}, \boldsymbol{e}_{t-1}\right).\end{aligned}$$

The objective function can be separated into two parts, and we take $C_t^{(1)}(\cdot)$ related to $z_t^{(1)}$ as an example to analyze the optimal APC, the other dimension $C_t^{(2)}$ is the same. Considering the randomness of parameters in decision making, we let:

$$\begin{aligned}C_t^{(1)}(\boldsymbol{u}_t) &= \mathbf{E}_{\hat{\theta}_{10}, \widehat{\boldsymbol{\theta}}_{\mathbf{1}}, \hat{\gamma}_1, \hat{\vartheta}_1, \hat{\omega}_1, \hat{\rho}_1, \hat{\varphi}_1, e_{t-1}^{(1)}}\left[z_t^{(1)}(\boldsymbol{u}_t) \middle| \hat{\theta}_{10}, \widehat{\boldsymbol{\theta}}_{\mathbf{1}}, \hat{\gamma}_1, \hat{\vartheta}_1, \hat{\omega}_1, \hat{\rho}_1, \hat{\varphi}_1, e_{t-1}^{(1)}\right]^2\\ &= \left(\mathbf{E}_{\hat{\theta}_{10}, \widehat{\boldsymbol{\theta}}_{\mathbf{1}}, \hat{\gamma}_1, \hat{\vartheta}_1, \hat{\omega}_1, \hat{\rho}_1, \hat{\varphi}_1, e_{t-1}^{(1)}}\left[z_t^{(1)}(\boldsymbol{u}_t) \middle| \hat{\theta}_{10}, \widehat{\boldsymbol{\theta}}_{\mathbf{1}}, \hat{\gamma}_1, \hat{\vartheta}_1, \hat{\omega}_1, \hat{\rho}_1, \hat{\varphi}_1, e_{t-1}^{(1)}\right]\right)^2\\ &\quad + \mathrm{var}_{\hat{\theta}_{10}, \widehat{\boldsymbol{\theta}}_{\mathbf{1}}, \hat{\gamma}_1, \hat{\vartheta}_1, \hat{\omega}_1, \hat{\rho}_1, \hat{\varphi}_1, e_{t-1}^{(1)}}\left[z_t^{(1)}(\boldsymbol{u}_t) \middle| \hat{\theta}_{10}, \widehat{\boldsymbol{\theta}}_{\mathbf{1}}, \hat{\gamma}_1, \hat{\vartheta}_1, \hat{\omega}_1, \hat{\rho}_1, \hat{\varphi}_1, e_{t-1}^{(1)}\right].\end{aligned} \tag{F.1}$$

Then we analyze these two items in Equation (F.1)

$$\mathbf{E}_{\hat{\theta}_{10},\hat{\boldsymbol{\theta}}_{\mathbf{1}},\hat{\gamma}_1,\hat{\vartheta}_1,\hat{\omega}_1,\hat{\rho}_1,\hat{\varphi}_1,e_{t-1}^{(1)}}[z_t^{(1)}(\boldsymbol{u}_t)|\hat{\theta}_{10},\hat{\boldsymbol{\theta}}_{\mathbf{1}},\hat{\gamma}_1,\hat{\vartheta}_1,\hat{\omega}_1,\hat{\rho}_1,\hat{\varphi}_1,e_{t-1}^{(1)}]$$
$$= E_{\hat{\theta}_{10},\hat{\boldsymbol{\theta}}_{\mathbf{1}},\hat{\gamma}_1,\hat{\vartheta}_1,\hat{\omega}_1,\hat{\rho}_1,\hat{\varphi}_1,e_{t-1}^{(1)}}[\theta_{10}+\theta_{11}u_t^{(1)}+\theta_{12}u_t^{(2)}+\theta_{13}u_t^{(3)}+\gamma_1 t+\vartheta_1 e_{t-1}^{(1)}+\omega_1 z_{t-1}^{(1)}+\varphi_1 t e_{t-1}^{(1)}+r_1|\hat{\theta}_{10},\hat{\boldsymbol{\theta}}_{\mathbf{1}},\hat{\gamma}_1,\hat{\vartheta}_1,\hat{\omega}_1,\hat{\rho}_1,\hat{\varphi}_1]$$
$$= \hat{\theta}_{10}+\hat{\theta}_{11}u_t^{(1)}+\hat{\theta}_{12}u_t^{(2)}+\hat{\theta}_{13}u_t^{(3)}+\hat{\gamma}_1 t+\hat{\vartheta}_1 e_{t-1}^{(1)}+\hat{\omega}_1 z_{t-1}^{(1)}+\hat{\varphi}_1 t e_{t-1}^{(1)}. \tag{F.2}$$

where $e_{t-1}^{(1)}$ denotes the regression noise of the first dimension for system output at run $t-1$. We denote the three dimensions of control recipe as $\boldsymbol{u}_t=\left[u_t^{(1)},u_t^{(2)},u_t^{(3)}\right]^{\boldsymbol{T}}$, and the corresponding parameters are $\boldsymbol{\theta}_{\mathbf{1}}=[\theta_{11},\theta_{12},\theta_{13}]$. By taking the conditional variance by the variable $e_{t-1}^{(1)}$, we have:

$$\begin{aligned}&\mathrm{var}_{\hat{\theta}_{10},\hat{\boldsymbol{\theta}}_{\mathbf{1}},\hat{\gamma}_1,\hat{\vartheta}_1,\hat{\omega}_1,\hat{\rho}_1,\hat{\varphi}_1,e_{t-1}^{(1)}}\left[z_t^{(1)}(\boldsymbol{u}_t)\middle|\hat{\theta}_{10},\hat{\boldsymbol{\theta}}_{\mathbf{1}},\hat{\gamma}_1,\hat{\vartheta}_1,\hat{\omega}_1,\hat{\rho}_1,\hat{\varphi}_1,e_{t-1}^{(1)}\right]\\&=\mathbf{E}_{e_{t-1}^{(1)}}\left[\mathrm{var}_{\hat{\theta}_{10},\hat{\boldsymbol{\theta}}_{\mathbf{1}},\hat{\gamma}_1,\hat{\vartheta}_1,\hat{\omega}_1,\hat{\rho}_1,\hat{\varphi}_1}\left[z_t^{(1)}(\boldsymbol{u}_t)\middle|\hat{\theta}_{10},\hat{\boldsymbol{\theta}}_{\mathbf{1}},\hat{\gamma}_1,\hat{\vartheta}_1,\hat{\omega}_1,\hat{\rho}_1,\hat{\varphi}_1,e_{t-1}^{(1)}\right]\right]\\&+\mathrm{var}_{e_{t-1}^{(1)}}\left[\mathbf{E}_{\hat{\theta}_{10},\hat{\boldsymbol{\theta}}_{\mathbf{1}},\hat{\gamma}_1,\hat{\vartheta}_1,\hat{\omega}_1,\hat{\rho}_1,\hat{\varphi}_1}\left[z_t^1(\boldsymbol{u}_t)\middle|\hat{\theta}_{10},\hat{\boldsymbol{\theta}}_{\mathbf{1}},\hat{\gamma}_1,\hat{\vartheta}_1,\hat{\omega}_1,\hat{\rho}_1,\hat{\varphi}_1,e_{t-1}^{(1)}\right]\right].\end{aligned} \tag{F.3}$$

For the first term in Equation (F.3), we have:

$$\mathbf{E}_{e_{t-1}^{(1)}}\left[\mathrm{var}_{\hat{\theta}_{10},\hat{\boldsymbol{\theta}}_{\mathbf{1}},\hat{\gamma}_1,\hat{\vartheta}_1,\hat{\omega}_1,\hat{\rho}_1,\hat{\varphi}_1}\left[z_t^{(1)}(\boldsymbol{u}_t)\middle|\hat{\theta}_{10},\hat{\boldsymbol{\theta}}_{\mathbf{1}},\hat{\gamma}_1,\hat{\vartheta}_1,\hat{\omega}_1,\hat{\rho}_1,\hat{\varphi}_1,e_{t-1}^{(1)}\right]\right]$$
$$=E_{e_{t-1}^{(1)}}\left[\mathrm{var}(\hat{\theta}_{10})+\boldsymbol{u}_t^T\boldsymbol{\Sigma}_{\boldsymbol{\theta}}^{\mathbf{1}}\boldsymbol{u}_t+t^2\mathrm{var}(\hat{\gamma}_1)+\left(e_{t-1}^{(1)}\right)^2\mathrm{var}(\hat{\vartheta}_1)+\left(z_{t-1}^{(1)}\right)^2\mathrm{var}(\hat{\omega}_1)+\mathrm{var}_{\hat{\varphi}_1}\left[\varphi_1 t e_{t-1}^{(1)}\right]+\mathrm{var}(r_1)\right]$$
$$=\mathrm{var}(\hat{\theta}_{10})+\boldsymbol{u}_t^T\boldsymbol{\Sigma}_{\boldsymbol{\theta}}^{\mathbf{1}}\boldsymbol{u}_t+t^2\mathrm{var}(\hat{\gamma}_1)+\left(e_{t-1}^{(1)}\right)^2\mathrm{var}(\hat{\vartheta}_1)+\left(z_{t-1}^{(1)}\right)^2\mathrm{var}(\hat{\omega}_1)+\mathrm{var}_{\hat{\varphi}_1}\left[\varphi_1 t e_{t-1}^{(1)}\right]+\mathrm{var}(r_1). \tag{F.4}$$

And the second term is:

$$\begin{aligned}&\mathrm{var}_{e_{t-1}^{(1)}}\left[\mathbf{E}_{\hat{\theta}_{10},\hat{\boldsymbol{\theta}}_{\mathbf{1}},\hat{\gamma}_1,\hat{\vartheta}_1,\hat{\omega}_1,\hat{\rho}_1,\hat{\varphi}_1}\left[z_t^1(\boldsymbol{u}_t)\middle|\hat{\theta}_{10},\hat{\boldsymbol{\theta}}_{\mathbf{1}},\hat{\gamma}_1,\hat{\vartheta}_1,\hat{\omega}_1,\hat{\rho}_1,\hat{\varphi}_1,e_{t-1}^{(1)}\right]\right]\\&=\mathrm{var}_{e_{t-1}^{(1)}}[\hat{\theta}_{10}+\hat{\theta}_{11}u_t^{(1)}+\hat{\theta}_{12}u_t^{(2)}+\hat{\theta}_{13}u_t^{(3)}+\hat{\gamma}_1 t+\hat{\vartheta}_1 e_{t-1}^{(1)}+\hat{\omega}_1 z_{t-1}^{(1)}+\hat{\varphi}_1 t e_{t-1}^{(1)}]\\&=\left(\hat{\vartheta}_1+\hat{\varphi}_1 t\right)^2\mathrm{var}\left(e_{t-1}^{(1)}\right).\end{aligned} \tag{F.5}$$

Therefore, we can summarize Equation (F.1) as

$$\begin{aligned}C_t^{(1)}(\boldsymbol{u}_t)=&[\hat{\theta}_{10}+\hat{\theta}_{11}u_t^{(1)}+\hat{\theta}_{12}u_t^{(2)}+\hat{\theta}_{13}u_t^{(3)}+\hat{\gamma}_1 t+\hat{\vartheta}_1 e_{t-1}^{(1)}+\hat{\omega}_1 z_{t-1}^{(1)}+\hat{\varphi}_1 t e_{t-1}^{(1)}]^2+\mathrm{var}(\hat{\theta}_{10})\\&+\boldsymbol{u}_t^T\boldsymbol{\Sigma}_{\boldsymbol{\theta}}^{\mathbf{1}}\boldsymbol{u}_t+t^2\mathrm{var}(\hat{\gamma}_1)+\left(e_{t-1}^{(1)}\right)^2\mathrm{var}(\hat{\vartheta}_1)+\left(z_{t-1}^{(1)}\right)^2\mathrm{var}(\hat{\omega}_1)+\mathrm{var}_{\hat{\varphi}_1}\left[\varphi_1 t e_{t-1}^{(1)}\right]\\&+\mathrm{var}(r_1)+\left(\hat{\vartheta}_1+\hat{\varphi}_1 t\right)^2\mathrm{var}\left(e_{t-1}^{(1)}\right).\end{aligned}$$

Similarly, for the second dimension, we have:

$$C_t^{(2)}(\boldsymbol{u}_t) = [\hat{\theta}_{20} + \hat{\theta}_{21}u_t^{(1)} + \hat{\theta}_{22}u_t^{(2)} + \hat{\theta}_{23}u_t^{(3)} + \hat{\gamma}_2 t + \hat{\vartheta}_2 e_{t-1}^{(2)} + \hat{\omega}_2 z_{t-1}^{(2)} + \hat{\varphi}_2 t e_{t-1}^{(2)}]^2 + \mathrm{var}(\hat{\theta}_{20})$$

$$+ \boldsymbol{u}_t^T \boldsymbol{\Sigma}_{\boldsymbol{\theta}}^{\mathbf{2}} \boldsymbol{u}_t + t^2 \mathrm{var}(\hat{\gamma}_2) + \left(e_{t-1}^{(2)}\right)^2 \mathrm{var}(\hat{\vartheta}_2) + \left(z_{t-1}^{(2)}\right)^2 \mathrm{var}(\hat{\omega}_2) + \mathrm{var}_{\hat{\varphi}_2}\left[\varphi_2 t e_{t-1}^{(2)}\right]$$

$$+ \mathrm{var}(r_2) + \left(\hat{\vartheta}_2 + \hat{\varphi}_2 t\right)^2 \mathrm{var}\left(e_{t-1}^{(2)}\right).$$

Then taking the first-order derivative of $C_t^{(1)}(\boldsymbol{u}_t) + C_t^{(2)}(\boldsymbol{u}_t)$, we have

$$\frac{\mathrm{d}\left(C_t^{(1)}(\boldsymbol{u}_t) + C_t^{(2)}(\boldsymbol{u}_t)\right)}{\mathrm{d}\boldsymbol{u}_t} = 2\left(\hat{\theta}_{10} + \hat{\boldsymbol{\theta}}_{\mathbf{1}}\boldsymbol{u}_t + \hat{\gamma}_1 t + \hat{\vartheta}_1 e_{t-1}^{(1)} + \hat{\omega}_1 z_{t-1}^{(1)} + \hat{\varphi}_1 t e_{t-1}^{(1)}\right)\hat{\boldsymbol{\theta}}_{\mathbf{1}} + 2\boldsymbol{u}_t \mathrm{var}(\hat{\boldsymbol{\theta}}_{\mathbf{1}}^{\boldsymbol{T}})$$

$$+2\left(\hat{\theta}_{20} + \hat{\boldsymbol{\theta}}_{\mathbf{2}}\boldsymbol{u}_t + \hat{\gamma}_2 t + \hat{\vartheta}_2 e_{t-1}^{(2)} + \hat{\omega}_2 z_{t-1}^{(2)} + \hat{\varphi}_2 t e_{t-1}^{(2)}\right)\hat{\boldsymbol{\theta}}_{\mathbf{2}} + 2\boldsymbol{u}_t \mathrm{var}(\hat{\boldsymbol{\theta}}_{\mathbf{2}}^{\boldsymbol{T}}) = 0.$$

If $\left[\Sigma_{\boldsymbol{\theta}}^{\mathbf{1}} + \hat{\boldsymbol{\theta}}_{\mathbf{1}}\hat{\boldsymbol{\theta}}_{\mathbf{1}}^{\boldsymbol{T}} + \Sigma_{\boldsymbol{\theta}}^{\mathbf{2}} + \hat{\boldsymbol{\theta}}_{\mathbf{2}}\hat{\boldsymbol{\theta}}_{\mathbf{2}}^{\boldsymbol{T}}\right]$ is invertible, we have the closed-form solution as:

$$\boldsymbol{u}_t^* = -\left[\Sigma_{\boldsymbol{\theta}}^{\mathbf{1}} + \hat{\boldsymbol{\theta}}_{\mathbf{1}}\hat{\boldsymbol{\theta}}_{\mathbf{1}}^{\boldsymbol{T}} + \Sigma_{\boldsymbol{\theta}}^{\mathbf{2}} + \hat{\boldsymbol{\theta}}_{\mathbf{2}}\hat{\boldsymbol{\theta}}_{\mathbf{2}}^{\boldsymbol{T}}\right]^{-\mathbf{1}} \cdot \left[\left(\hat{\theta}_{10} + \hat{\gamma}_1 t + \hat{\vartheta}_1 e_{t-1}^{(1)} + \hat{\varphi}_1 t e_{t-1}^{(1)} + \hat{\omega}_1 z_{t-1}^{(1)}\right) \cdot \hat{\boldsymbol{\theta}}_{\mathbf{1}} + \left(\hat{\theta}_{20} + \hat{\gamma}_2 t + \hat{\vartheta}_2 e_{t-1}^{(2)} + \hat{\varphi}_2 t e_{t-1}^{(2)} + \hat{\omega}_2 z_{t-1}^{(2)}\right) \cdot \hat{\boldsymbol{\theta}}_{\mathbf{2}}\right].$$